\documentclass[10pt,twocolumn,letterpaper]{article}

\usepackage[pagenumbers]{paper}   % 

\usepackage{graphicx}
\usepackage{booktabs}
\usepackage[dvipsnames]{xcolor}
\usepackage{makecell}
\usepackage{multirow}
\usepackage{multirow}

\usepackage{multirow}
\usepackage{arydshln}
\definecolor{darkblue}{HTML}{00008B}
\definecolor{lightblue}{HTML}{ADD8E6}
\definecolor{rred}{HTML}{d43e4f}
\definecolor{ppurple}{HTML}{5e4fa2}
\definecolor{ccyan}{HTML}{67c1a5}
\definecolor{ggreen}{HTML}{acdda5}
\definecolor{llime}{HTML}{e6f598}
\definecolor{yyellow}{HTML}{ffffbf}
\definecolor{llightorange}{HTML}{fdae61}
\definecolor{bblue}{HTML}{3287bd}
\definecolor{oorange}{HTML}{f46e43}
\definecolor{rrred}{HTML}{e27884}
\definecolor{pppurple}{HTML}{8f84be}
\definecolor{cccyan}{HTML}{94d3be}
\definecolor{gggreen}{HTML}{c3e6be}
\definecolor{lllime}{HTML}{edf8b6}
\definecolor{yyyellow}{HTML}{ffffd3}
\definecolor{lllightorange}{HTML}{fec590}
\definecolor{bbblue}{HTML}{a8cce2}
\definecolor{ooorange}{HTML}{f7997d}
\definecolor{hard}{HTML}{A9A9A9}
\definecolor{simple}{HTML}{D3D3D3}
\definecolor{lightblue}{RGB}{173, 216, 230}
\definecolor{lightgreen}{RGB}{144, 238, 144}
\definecolor{lightyellow}{RGB}{255, 255, 224}
\definecolor{lightred}{RGB}{255, 182, 193}
\definecolor{lightpurple}{RGB}{216, 191, 216}
\definecolor{darkblue}{RGB}{123, 166, 180}   
\definecolor{darkgreen}{RGB}{94, 188, 94}    
\definecolor{darkyellow}{RGB}{245, 245, 184} 
\definecolor{darkred}{RGB}{205, 132, 143}    
\definecolor{darkpurple}{RGB}{166, 141, 166} 
\usepackage{arydshln}
\usepackage{makecell}

\newcommand{\xmark}{\ding{55}}

\usepackage{pifont}
\renewcommand{\xmark}{\ding{55}} % ✗

\usepackage{tcolorbox}
\tcbuselibrary{breakable}
\usepackage{algorithm}
\usepackage{algorithmic}
\definecolor{paperblue}{rgb}{0.21,0.49,0.74}
\usepackage[pagebackref,breaklinks,colorlinks,allcolors=paperblue]{hyperref}
\def\paperID{ } % *** Enter the Paper ID here
\def\confName{paper}
\def\confYear{2026}

\title{UniBYD: A Unified Framework for Learning Robotic Manipulation Across Embodiments \textbf{\textcolor{Maroon}{B}}e\textbf{\textcolor{Maroon}{y}}on\textbf{\textcolor{Maroon}{d}} Imitation of Human Demonstrations}

\author{
    Tingyu Yuan$^{1,2}$ \quad
    Biaoliang Guan$^{3}$ \quad
    Wen Ye$^{1,2}$ \quad
    Ziyan Tian$^{1,2}$ 
    Yi Yang$^{4}$ \quad 
    Weijie Zhou$^{5}$ \quad \\
    Zhaowen Li$^{6}$ \quad
    Yan Huang$^{1,2}$ \quad 
    Peng Wang$^{1,2}$ \quad
    Chaoyang Zhao$^{1 \dagger}$ \quad
    Jinqiao Wang$^{1,2 \dagger}$ \quad 
    \vspace{0.3em} \\ % 增加一点垂直间距
    $^{1}$CASIA \quad
    $^{2}$UCAS  \quad
    $^{3}$XJTU \quad
    $^{4}$CSU \quad
    $^{5}$BJTU \quad 
    $^{6}$Yinwang Intelligent Technology Co. Ltd. \quad
    \vspace{0.3em} \\ % 增加一点垂直间距
}

\begin{document}
\maketitle
\begingroup
    \renewcommand\thefootnote{} % 清除脚注编号（防止出现数字）
    \footnotetext{$^\dagger$Corresponding authors.}
\endgroup
\begin{abstract}
  In embodied intelligence, the embodiment gap between robotic and human hands brings significant challenges for learning from human demonstrations. 
Although some studies have attempted to bridge this gap using reinforcement learning, they remain confined to merely reproducing human manipulation, resulting in limited task performance. Moreover, current methods struggle to support diverse robotic hand configurations. In this paper, we propose UniBYD, a unified framework that uses a dynamic reinforcement learning algorithm to discover manipulation policies aligned with the robot’s physical characteristics. To enable consistent modeling across diverse robotic hand morphologies, UniBYD incorporates a unified morphological representation (UMR).
Building on UMR, we design a dynamic PPO with an annealed reward schedule, enabling reinforcement learning to transition from offline-informed imitation of human demonstrations to online-adaptive exploration of policies better adapted to diverse robotic morphologies, thereby going beyond mere imitation of human hands.
To address the severe state drift caused by the incapacity of early-stage policies, we design a hybrid Markov-based \textit{shadow engine} that provides fine-grained guidance to anchor the imitation within the expert's manifold.
To evaluate UniBYD, we propose UniManip, the first benchmark for cross-embodiment manipulation spanning diverse robotic morphologies. Experiments demonstrate a 44.08\% average improvement in success rate over the current state-of-the-art. Our project page is \url{https://zhanheng-creator.github.io/UniBYD/}.
\end{abstract} 
    
\section{Introduction}
\label{sec:intro}
Artificial intelligence has advanced rapidly in recent years \cite{yang2026focus, yuan2026ideavatar, yang2026tracevision}, with remarkable progress in robotics. Within embodied intelligence, learning from human demonstrations \cite{zhou2025you, ye2024latent, qin2022dexmv, luo2025being,liu2024quasisim} has emerged as a dominant paradigm. However, the embodiment gap between human hands and robotic hands of varying morphologies \cite{lum2025crossing} poses significant challenges for this area, which existing studies have yet to effectively address. Notably, most research focuses on anthropomorphic hands, while cross-embodiment generalization to 2/3-fingered grippers remains largely unexplored. For example, retargeting-based methods typically map only the kinematic poses while ignoring critical dynamic information. Existing imitation learning methods remain at merely reproducing human operations\cite{wu2023learning,qin2022dexmv}. Given the morphological and dynamic discrepancies between human hands and robot hands of varying structures, such as differences in finger count and degrees of freedom, such direct reproduction limits them far below the level demonstrated by humans.

\begin{figure}[t]
  \centering
  \includegraphics[width=\linewidth]{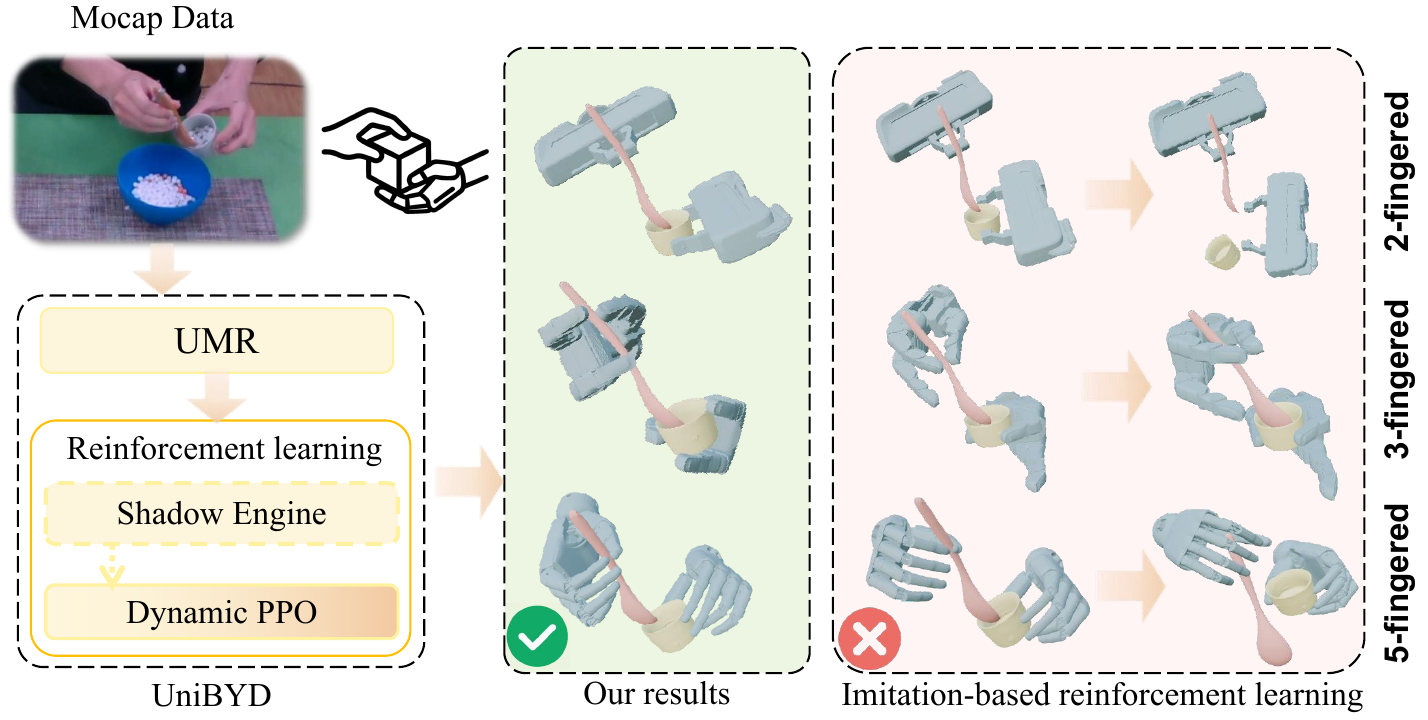}
   \caption{Leveraging human demonstrations, UniBYD learns manipulation strategies that transcend mere imitation and are tailored to a broad spectrum of robotic hand morphologies.}
   \label{fig:1}
\end{figure}

To overcome the limitations of the above approaches, researchers have begun exploring reinforcement learning from human data. However, such methods often struggle to discover strategies that are truly aligned with the robot’s own morphology, resulting in suboptimal task performance, especially for non-anthropomorphic hands with significant topological deviations. 
Some studies, such as ManipTrans \cite{li2025maniptrans}, employ Proximal Policy Optimization (PPO) \cite{schulman2017proximal} in simulated environments, yet their reward functions merely enforce strict time alignment of the robot hand’s joint angles with the expert trajectory at every step, which only partially addresses the inherent limitations of imitation learning. As shown on the right of \cref{fig:1}, such methods merely map human-hand motions onto robotic hands, resulting in low success rates for task execution.
Secondly, another line of research attempts to completely eliminate dependence on human demonstrations by defining reward functions centered on object pose errors \cite{chen2024object}. This comes at the cost of losing the crucial guidance that human prior provides, making it difficult to approach the high-performance policy regions. These methods tend to fall into local optima.
Beyond these limitations, a fundamental challenge for reinforcement learning in this domain is the severe state drift in early training. Due to the lack of effective guidance, even minor deviations from human priors can cause the robot to stray from correct trajectories and terminate episodes prematurely, resulting in extremely limited information gain and failed bootstrapping.
Moreover, existing methods exhibit limited generalization, as most are tailored to specific robotic hands and lack a unified framework for adapting human policies to diverse robots \cite{yuan2025hermes}.

To address these challenges, we introduce a novel paradigm that shifts from rigid motion imitation to morphology-adaptive policy discovery, ensuring manipulation strategies are inherently aligned with the mechanical characteristics of diverse robotic embodiments.
According to this paradigm, we introduce UniBYD, a unified reinforcement learning framework designed to acquire morphology-adaptive manipulation policies that transcend mere imitation and generalize across diverse embodiments.
Firstly, to realize cross-embodiment generalization, we introduce a unified morphological representation (UMR), which standardizes the state-action space and employs morphology embeddings to enable configuration-aware modeling.
Building on UMR, we propose a dynamic PPO mechanism with reward annealing, which enables a smooth transition from offline-informed imitation to online-adaptive exploration. This process guides the model to discover policies better suited to the physical potential of diverse robots.
To mitigate the severe state drift in early training, we design a hybrid Markov-based \textit{shadow engine} that provides fine-grained guidance to anchor the robot's execution near expert trajectories, ensuring consistent information gain.

To our knowledge, UniBYD is the first to learn manipulation policies for diverse embodiments from human demonstrations with reinforcement learning. To rigorously evaluate UniBYD, we build UniManip, the first benchmark that spans diverse hand configurations and a wide variety of tasks, providing a standardized measure for manipulation competence. Experimental results show that UniBYD achieves a substantial improvement of up to 44.08\% in overall success rate over the current state-of-the-art (SOTA). Our contributions are as follows:
\begin{itemize}
    \item  We propose UniBYD, a unified reinforcement learning framework compatible with various robotic hand types. The framework learns control strategies aligned with the diverse embodiments better.
    \item We design a dynamic PPO, which facilitates a systematic transition from offline-informed imitation to online-adaptive exploration, enabled by a hybrid Markov-based \textit{shadow engine} that anchors early-stage training to prevent trajectory drift.
    \item We construct UniManip, the first unified benchmark based on human demonstrations for evaluating robotic manipulation capability across morphologies. Experiments on UniManip demonstrate that UniBYD significantly outperforms existing SOTA methods, achieving a 44.08\% gain in success rate.
\end{itemize}

\section{Related Work}
\label{sec:formatting}
\textbf{Robotic manipulation learning methods.} Achieving human-level dexterity remains a central challenge in robotics \cite{song2025overview, li2025developments, zhong2021extending, jin2024complementarity, gao2023real, gao2024enhancing, yang2025multi, zhao2023gelsight, manawadu2024dexterous, huang2025dexterous, luo2024precise}. Classical approaches such as trajectory optimization \cite{suh2025dexterous, patel2019contact, yang2025multi} and STOCS \cite{zhang2025simultaneous} compute joint trajectories and contact forces, yet are predominantly offline and thus limited in real-time applicability. Model Predictive Control \cite{ai2025review, jiang2024contact, jin2024complementarity} enables online planning via finite-horizon optimization, but its computational burden hampers real-time deployment. Reinforcement learning optimizes policies through environment interaction \cite{yin2023rotating,cutler2024benchmarking}, supporting the training of complex tasks. However, the exploration space is vast, a challenge UniBYD addresses by using human demonstrations to provide crucial guidance and rapidly focus the policy search.

\noindent \textbf{Learning from Human Demonstrations.} Given the high cost of collecting robotic manipulation data, numerous methods have sought to learn from human demonstrations \cite{rajeswaran2017learning, zhu2019dexterous}. Conventional inverse-kinematics retargeting \cite{lin2025dexflow, xin2025analyzing} and learning-based retargeting \cite{lin2025dexflow, shaw2024learning, park2025learning} offer limited performance. Mainstream practice leverages reinforcement learning to acquire robotic manipulation skills from human hand data. Imitation-centric approaches map demonstrations directly onto robots \cite{qin2022dexmv}. For example, ManipTrans \cite{li2025maniptrans} reproduces human actions on a 5-fingered platform via a universal trajectory imitator and residual modules. However, it fails to surpass the imitation of demonstrations and ignores the embodiment gap. In contrast, goal-centric methods\cite{lum2025crossing}, such as DexMachina \cite{mandi2025dexmachina},  prioritize task objectives while largely ignoring imitation rewards to encourage exploration, but training is lengthy and convergence is often precarious.
Moreover, developing a unified framework that generalizes across diverse robotic hand morphologies remains a critical challenge.

\begin{figure*}[t]
  \centering
  \includegraphics[width=\textwidth]{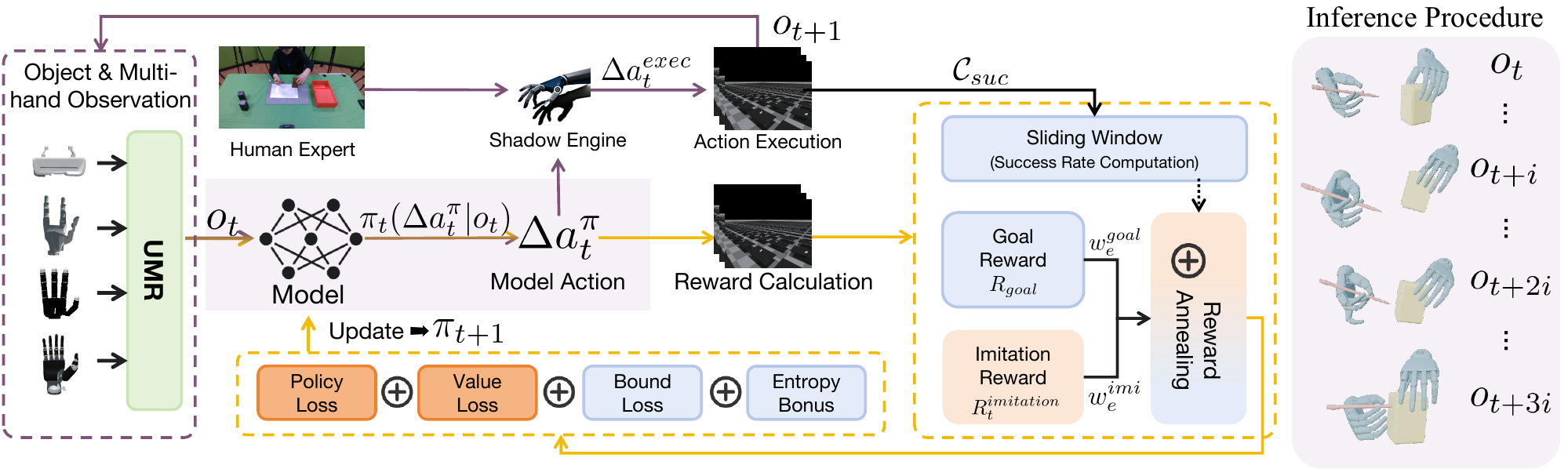}
   \caption{The framework of UniBYD. UniBYD first encodes diverse hands via UMR. It then employs a dynamic PPO with an annealed reward mechanism, which initially leverages the \textit{shadow engine} for high-fidelity imitation before transitioning to autonomous exploration to discover morphology-aligned policies.}
   
   \label{fig:fig2}
\end{figure*}

\noindent \textbf{Benchmarks.} Existing benchmarks for robotic manipulation have been proposed \cite{li2025teleopbench}, including Bi-DexHands \cite{chen2023bi}, which assembles dozens of bimanual tasks in simulation, and VTDexManip \cite{liuvtdexmanip}, which evaluates single- and bimanual manipulation with integrated vision and touch. Benchmarks tailored to LfHD have likewise emerged \cite{ye2025dex1b}; for instance, EgoDex \cite{hoque2025egodex} compiles 829 hours of egocentric manipulation videos and hand-motion data, and DexMachina \cite{mandi2025dexmachina} introduces a benchmark centered on articulated objects. Nevertheless, a comprehensive, unified benchmark spanning 2-, 3-, and 5-fingered hand morphologies across diverse single-hand and bimanual tasks is still lacking. Moreover, existing evaluation protocols are largely one-dimensional, failing to assess manipulation strategies and their alignment with diverse robotic hand embodiments.

\section{Method}
As shown in \cref{fig:fig2}, we propose UniBYD, a unified and progressive reinforcement learning framework that learns from humans across various robotic hands. 
UniBYD aims to discover morphology-aligned policies that transcend mere imitation. 
During the early training phase, UniBYD blends the action $\Delta a_t^{\pi}$ predicted by the policy network with the expert action $\Delta a_t^{E}$ from the demonstration to generate the final executed action used to advance the environment to the next state. In contrast, during the inference phase, UniBYD relies solely on the policy network, directly executing its predicted action $\Delta a_t^{\pi}$ to complete the task.

\subsection{Unified Morphological Representation}
The challenge of cross-morphology generalization lies in the fundamental differences among robotic hand embodiments. Accordingly, we propose an efficient unified morphological representation.

For robotic hand $h$, the proprioceptive state $s_t^{h}$ at step $t$ includes a fixed-dimensional wrist state $s_{base} \in \mathbb{R}^{13}$ and a variable-dimensional joint state $s_{joint}^{h}$. The wrist state encodes position, orientation, and velocities, while the joint state contains joint angles and velocities, $q^{h}, \dot{q}^{h} \in \mathbb{R}^{D_h}$, where $D_h$ is the hand’s degrees of freedom. To avoid the $2\pi$ wrap-around issue, joint angles are trigonometrically encoded as $\cos(q^{h})$ and $\sin(q^{h})$.

Let $D_{max}$ denote the maximum number of joint degrees of freedom. For hands with $D_h < D_{max}$, we apply zero-padding to the variable component $s_{joint}^{h}$, thereby elevating its dimensionality to $D_{max}$ and obtaining $s_{joint}^{pad}$:
\begin{equation}
\begin{split}
  s_{joint}^{pad} = [ & q^{h} \oplus \mathbf{0}_{D_{max} - D_h} \oplus \cos(q^{h}) \oplus \mathbf{0}_{D_{max} - D_h} \oplus \\
  & \sin(q^{h}) \oplus \mathbf{0}_{D_{max} - D_h} \oplus \dot{q}^{h} \oplus \mathbf{0}_{D_{max} - D_h} ].
\end{split}
\label{eq:important}
\end{equation}

Moreover, the policy $\pi_\theta$ must be informed of the hand’s specific physical attributes. To this end, we extract key static morphological properties from the hand’s URDF model, namely $D_h$, the number of fingers $N_{finger}^{h}$, and the number of rigid bodies $N_{body}^{h}$. These quantities constitute a static descriptor $v_{morph}^{h} = [ N_{finger}^{h}, D_h, N_{body}^{h} ]$.

Finally, we concatenate $s_{base}$, $s_{joint}^{pad}$, and $v_{morph}^{h}$ to form the observation $o_t^{finger} = s_{base} \oplus s_{joint}^{pad} \oplus v_{morph}^{h}$.
By unifying dynamic states and static attributes into a fixed-dimensional representation, UMR enables the policy to adapt to diverse hand morphologies and learn morphology-specific manipulation policies.

\subsection{Dynamic Proximal Policy Optimization}

With UMR providing a consistent observation space, UniBYD employs a reinforcement learning algorithm integrating a reward annealing mechanism. This mechanism facilitates a systematic transition from offline-informed imitation to online-adaptive exploration, allowing the model to inherit expert knowledge while transcending its limitations through active environmental interaction.

\subsubsection{From Offline-Informed Imitation to Online-Adaptive Exploration}

\textbf{ (1) Imitation Reward ($R^{imitation}$).} To achieve precise imitation of offline expert demonstrations, we design a dense, multi-component imitation reward $R^{imitation}$ that, at step $t$ of episode $i$, quantifies the similarity between the current state $s_t$ and the expert state $s_t^E$. It is defined as
\begin{equation}
R_t^{imitation} = \sum_{k=1}^{n} w_k \cdot r_k(s_t, s_t^E) - p(\Delta a_t),
\label{eq:important}
\end{equation}
where $r_k$ denotes the $k$-th reward component and $w_k$ its corresponding weight. The components primarily encompass discrepancies in wrist pose, linear and angular velocities; fingertip positions and contact forces; joint positions and velocities; and object pose, linear and angular velocities. The term $p(\Delta a_t)$ is a power penalty on actuation. For detailed computations, see Appendix Section 7.

\textbf{(2) Goal Reward ($R^{goal}$).} To relax the reliance on offline expert demonstrations in dexterous manipulation, we define a sparse goal reward $R^{goal}$ that depends solely on task success. In contrast to the dense $R_t^{imitation}$, this signal is step-agnostic and granted only when the entire episode is successfully completed, thereby conferring a substantial bonus that marks the attempt as productive exploration. $R^{goal}$ is defined as:
\begin{equation}
R^{goal} =
\begin{cases}
S_{bonus} & \text{if } \mathcal{C}_{suc} = \text{1}  \\
0 & \text{otherwise}
\end{cases},
\label{eq:important}
\end{equation}
where $S_{bonus}$ is a fixed-magnitude bonus, and $\mathcal{C}_{suc} \in \{0, 1\}$ denotes the episodic success indicator. In the first phase, $\mathcal{C}_{suc}$ is determined by the discrepancies between the object and fingertip observations and those of the expert, and later solely by the object. More details are provided in Appendix Section 8.

\textbf{(3) Dynamic Reward Annealing.} To enable a smooth transition from imitation to task-centric exploration, we define the total reward $R_t$ as a dynamically weighted sum of the imitation reward $R_t^{imitation}$ and the goal reward $R^{goal}$:
\begin{equation}
R_t =w_e^{imi} \cdot R_t^{imitation} + w_e^{goal} \cdot R^{goal},
\label{eq:important}
\end{equation}
where $w_e^{imi}$ and $w_e^{goal}$ are weighting coefficients for $R_t^{imitation}$ and $R^{goal}$ at training epoch $e$. The evolution of these weights constitutes a three-stage curriculum, with phase transitions jointly governed by the epoch threshold $T_{decay}$  marking the end of the \textit{shadow engine}, the earliest trigger threshold $T_{early}$, the recent success rate $\bar{SR}$, and a scaling factor $\gamma_{scale}$ controlling the transition speed.

To evaluate the current competence of the model in real time, we introduce the recent success rate $\bar{SR}$ as a performance indicator. Specifically, we design a sliding window of size $M$ to aggregate the outcomes of the most recent $M$ episodes, defined as $\bar{SR} = \frac{1}{M} \sum_{j=i-M+1}^{i} \mathcal{C}_{suc}$.

To quantify the progression of task mastery, we introduce a minimal competence threshold $\delta_{min}$ (set to $0.2$) and a transition factor $f$:
\begin{equation}
f = \max\left(0, \min\left(1, (\bar{SR} - \delta_{min}) \times \gamma_{scale}\right)\right).
\end{equation}
The term $(\bar{SR} - \delta_{min})$ primarily serves to prevent $w_e^{imi}$ from decaying under sub-optimal performance or stochastic noise. Simultaneously, it mitigates abrupt weight drops upon triggering the transition mechanism, thereby maintaining training stability during the switchover phase.

Furthermore, we introduce a success-rate trigger threshold $\delta_{SR}$ as a criterion to determine whether the model has attained sufficient proficiency to initiate the transition. Building on this, the weights $w_e^{imi}$ and $w_e^{goal}$ are adaptively adjusted according to the following formulations:
\begin{equation}
\resizebox{0.9\linewidth}{!}{$
w_e^{imi} = \begin{cases}
w_{max}^{imi}, \quad \text{if } e \le T_{decay} \land \\
               \quad \quad\quad\quad (e \le T_{early} \lor \bar{SR} \le \delta_{SR}) \\
\max(w_{min}^{imi}, w_{max}^{imi} \times (1.0 - f)), \quad \text{otherwise}
\end{cases},
$}
\end{equation}

\begin{equation}
\resizebox{0.85\linewidth}{!}{$
w_e^{goal} = \begin{cases}
0, \quad 
     \quad\text{if } e \le T_{decay} \land \\
     \quad\quad\quad (e \le T_{early} \lor \bar{SR} \le \delta_{SR})\\
\max(w_{min}^{goal}, w_{max}^{goal} \times f), \quad \text{otherwise}
\end{cases}.
$}
\end{equation}

This dynamic reward annealing approach implicitly partitions the training process into three stages. In the early imitation-driven alignment phase, while the initial temporal or performance criteria are not yet met, the weights remain fixed to prioritize learning from expert demonstrations and establishing basic manipulation skills. During the hybrid adaptive transition phase, if the model’s competence remains limited ($\bar{SR} - \delta_{min} \le 0$), $w_e^{imi}$ is maintained without decay while $w_e^{goal}$ is set to a small value to introduce $R^{goal}$. Once the model has mastered the manipulation skills, $w_e^{imi}$ enters an annealing mode, decreasing gradually according to changes in $\bar{SR}$ as $w_e^{goal}$ increases. At this stage, the influence of expert data begins to wane, facilitating a "soft-handover" of the guidance focus toward $R^{goal}$. Finally, in the autonomous exploration phase, the imitation weight reaches its minimum $w_{min}^{imi}$, exerting a nearly negligible influence. At this point, the policy is predominantly driven by the sparse goal reward, focusing solely on the overall success of the task. This allows the model to move beyond strictly mirroring human manipulation and discover refined control strategies that are better optimized for the robot hand's unique hardware morphology.

\subsubsection{Loss Synergy and Counterbalancing}

To facilitate effective exploration and prevent premature convergence, we incorporate the entropy regularization and boundary loss into the PPO objective, forming a synergy–counterbalance strategy that ensures policy physical feasibility while pursuing optimal performance.

\textbf{(1) Entropy Regularization.} 
To prevent the policy, particularly during the imitation-guided reinforcement learning phase, from prematurely converging to a suboptimal deterministic solution, we introduce an entropy regularization term, $\mathcal{H}(\pi_{\theta}(\cdot \mid o_t))$. This term serves as an entropy bonus that encourages sustained exploration. Its coefficient, $c_e^{\text{entropy}}$, follows a linear decay schedule:
\begin{equation}
c_e^{\text{entropy}} = \max\!\left(0,\; c_{\text{start}}^{\text{entropy}} \left(1 - \frac{e}{T_{\text{entropy\_decay}}}\right)\right),
\label{eq:important}
\end{equation}
where $c_{\text{start}}^{\text{entropy}}$ is the initial entropy coefficient, and $T_{\text{entropy\_decay}}$ is the decay horizon. This mechanism ensures ample exploration in the early stages of training and gradually reduces exploration thereafter to facilitate convergence.

\textbf{(2) Bound Loss.} While the entropy bonus fosters exploration, the mean of the policy distribution, $\mu_{\theta}(o_t)$, can readily drift beyond the physical action space. Conventional hard clipping disrupts gradient flow and severely impairs training. To remedy this, we introduce a differentiable soft boundary loss,  $L^{bound}$, which penalizes only clearly out-of-bounds means $\mu_t$:
\begin{equation}
\begin{split}
  L_t^{bound}(\mu_t) = [ & \sum_{j=1}^{D_a} ( \max(0, \mu_{t,j} - \mu_{bound})^2 + \\
  & \max(0, -\mu_{t,j} - \mu_{bound})^2 )],
\end{split}
\label{eq:important}
\end{equation}
where $D_a$ denotes the action dimensionality, $\mu_{t,j}$ is the j-th component of the action-mean vector $\mu_t$ , and $\mu_{bound}$ is a soft boundary threshold. 

\textbf{(3) Dynamic PPO Objective Function.} 
We integrate the foregoing components into the PPO objective, which we minimize:
\begin{equation}
\begin{split}
  L_t(\theta) = [ & -L_t^{\text{CLIP}}(\theta) + c_{vf} L_t^{\text{VF}}(\theta) - \\
  & c_e^{\text{entropy}} \mathcal{H}(\pi_{\theta}(\cdot \mid o_t)) + c_{\text{bound}} L_t^{\text{bound}}(\mu_t)],
\end{split}
\label{eq:important}
\end{equation}
where $L_t^{\text{CLIP}}$ denotes PPO’s clipped surrogate objective, $L_t^{\text{VF}}$ is the mean-squared error loss of the value function, and $c_{vf}$ and $c_{\text{bound}}$ are their respective weighting coefficients. The entropy bonus term ($- c_e^{\text{entropy}} \mathcal{H}$) and the boundary loss term ($+ c_{\text{bound}} L_t^{\text{bound}}$) establish an effective synergy-and-counterbalance: the former fosters broad exploration, while the latter ensures that such exploration remains confined to a physically safe and smooth action space.

\subsection{Hybrid Markov-based Shadow Engine}
\begin{figure}[t]
  \centering
  \includegraphics[width=\linewidth]{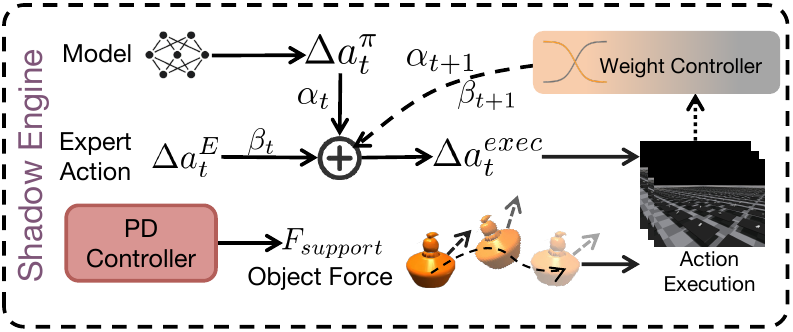}
   \caption{Overview of action generation and object control in the \textit{shadow engine}. It blends the model-predicted action and the expert-guided action to generate the final executed action $\Delta a_t^{exec}$. A PD controller applies an expert object force to guide the object.}
   \label{fig:fig3}
\end{figure}

At the outset of training, the policy network $\pi_\theta$ is markedly weak. Within a standard Markov Decision Process (MDP), even a slight action deviation $\Delta a_t$ can shift the subsequent state $s_{t+1}$, and the ensuing compounding errors rapidly drive the policy away from meaningful expert trajectories.
However, it is difficult for the policy to return to the correct path by relying on the penalties provided by post-hoc metrics $R^{imitation}$.
As a result, this leads to frequent premature episode terminations. Ultimately, the model receives only scarce and weak training signals, impairing overall learning efficiency.
Therefore, UniBYD introduces the \textit{shadow engine} to guide the dynamic PPO during the crucial early phase as shown in \cref{fig:fig3}. It enables fine-grained, efficient imitation by applying dynamic expert guidance to both the robotic hands and the object.

\subsubsection{Dexterous Hand Control}
To mitigate compounding errors caused by an initially weak policy, we propose a Hybrid MDP mechanism that unifies discrete pointwise learning with continuous Markov process learning. Specifically, at training step $t$, the action executed in the simulator, $\Delta a_t^{exec}$, is not the raw prediction $\Delta a_t^{\pi}$ from $\pi_\theta$. It is a dynamically weighted blend of $\Delta a_t^{\pi}$ and the expert demonstration action $\Delta a_t^{E}$.

At step $t$, conditioned on the current observation $o_t$, the policy network produces the action $\Delta a_t^{\pi}$. Concurrently, we retrieve the corresponding expert action $\Delta a_t^{E}$ from the demonstration. The executed action $\Delta a_t^{exec}$ is defined as
\begin{equation}
\Delta a_t^{exec} = \alpha_t \cdot \Delta a_t^{\pi} + \beta_t \cdot \Delta a_t^{E},
\label{eq:important}
\end{equation}
where $\alpha_t$ is the weight on the policy action  $\Delta a_t^{\pi}$, and $\beta_t$ is the weight on the expert action $\Delta a_t^{E}$. These weights satisfy $\alpha_t + \beta_t = 1$ throughout training.

We realize dynamic guidance by applying a linear decay curriculum to $\beta_t$ as a function of the training epoch $e$:
\begin{equation}
\beta_t = \max\left(0, 1 - \frac{e}{T_{decay}}\right),
\label{eq:important}
\end{equation}
where $T_{decay}$  is a predefined decay horizon. Correspondingly, $\alpha_t$ increases linearly from 0 to 1. This setup ensures that in the early phase of training $\beta_t \approx 1$ and $\alpha_t \approx 0$, so the model effectively learns each step in isolation without being influenced by the previous step. Once imitation performance is high, $\beta_t = 0$ and $\alpha_t = 1$. At that point $a_t^{exec} = a_t^{\pi}$, the guidance of the \textit{shadow engine} disappears, and $\pi_\theta$ must independently handle the full Markov decision process. See Appendix Section 6.1 for details.

\subsubsection{Object Control}
In complex tasks, the object's intrinsic physical properties, such as gravity and inertia, can still precipitate failure. To further ease the initial learning phase, the \textit{shadow engine} also applies a dynamic support force $F_{support}$ directly to the object. We utilize a proportional-derivative (PD) controller to compute $F_{support}$ based on the desired object pose and velocity ($g_t^{obj}, \dot{g}_t^{obj}$) from the expert demonstration.
Acting as an invisible hand, $F_{support}$ constrains the object to remain near its target trajectory, preventing drops or catastrophic deviations. The gains of this PD controller ($K_p, K_d$) are also gradually decayed to zero as training proceeds. For further details, see Appendix Section 6.2.
\section{Experiments}
\subsection{Benchmark}
To evaluate generalizable manipulation across diverse hand morphologies, a benchmark must encompass a wide spectrum of tasks involving complex object interactions and diverse functional manipulations. 
We build UniManip upon OakInk-V2 \cite{zhan2024oakink2}, as it represents one of the largest-scale and most diverse human demonstration datasets, featuring unrivaled synchronization precision and task complexity.
Leveraging these high-fidelity motion data, we curate a wide range of unimanual and bimanual tasks and convert them into expert demonstrations tailored to robotic hands with 2, 3, or 5 fingers (see Appendix Section 10 for conversion details and task taxonomy).
In total, we assemble 31 task categories to construct UniManip. For 5-finger unimanual and bimanual settings, we select 9 and 6 task categories, respectively. Given that 3-finger and 2-finger hands are ill-suited to bimanual manipulation, we evaluate them on 8 unimanual task categories each. We assess manipulation performance along multiple axes, and specifically define the following metrics:
\begin{itemize}
    \item  \textbf{Position Error (PE):} PE is defined as the mean Euclidean distance between the observed and target object positions across time steps in successful episodes. If all test episodes fail, PE is assigned a default value of 3. Lower PE indicates more accurate and stable manipulation.
    \item \textbf{Orientation Error (OE):} OE is defined as the mean minimal rotation angle between the observed and target orientations across time steps in successful episodes. If all test episodes fail, OE is set to a default value of 30. Lower OE reflects more precise orientation control.
    \item \textbf{Success Rate (SR):} The percentage of test episodes classified as successful. Given the process-oriented nature of our tasks, an episode is deemed successful if and only if every time step simultaneously satisfies PE $\le$ 3 cm and OE $\le$ 30°. Any violation at any time step renders the episode a failure.
    \item \textbf{Adaptation Score (AS):} This metric aims to quantify how well a manipulation policy aligns with the robot’s hardware morphology beyond binary success. We employ a joint evaluation framework, incorporating a suite of heterogeneous large models (including Gemini 2.5 Pro, GPT-5.2, and Qwen2.5-VL-72B) alongside 100 human volunteers as expert evaluators to provide a composite score from 0 to 10. LLMs and humans are blind to success labels to ensure independent assessment of manipulation quality.
    Please refer to Appendix Section 10.3 for calculation details and prompts. Crucially, AS serves as a qualitative complement to our primary physical metric.
\end{itemize}

\subsection{Experimental Setup}
\textbf{Implementation details.} Simulations were conducted in Isaac Gym\cite{makoviychuk2021isaac}, with 4,096 parallel environments during training. We configured $S_{bonus}$ to 20, $ T_{decay}$ to 150, $T_{early}$ to 100, $\gamma_{scale}$ to 2, $\delta_{min}$ to 0.2, and $\delta_{SR}$ to 0.8. Additional parameters are detailed in Appendix Section 9.1. For each task in UniManip, we executed 1,000 trials per task in simulation.

\begin{table}[t]
    \small

    \centering
    \caption{Comparative results on UniManip. UniBYD consistently outperforms representative baselines across diverse hand morphologies.}
    \label{tab:1}
    \resizebox{\linewidth}{!}{
        \setlength{\tabcolsep}{2pt} 
        \begin{tabular}{@{}llcccc@{}} 
        \toprule
        \makecell{\textbf{Hand} \\ \textbf{Type}} &
        \textbf{Metrics} &
        \makecell{\textbf{Reta} \\ \textbf{rget}} &
        \makecell{\textbf{Manip} \\ \textbf{trans}} &
        \makecell{\textbf{DexMa} \\ \textbf{china$^*$}} &
        \textbf{UniBYD} \\ 
        \midrule
        \multirow{4}{*}{\makecell{2 (1 \\ hand)}} 
        & SR$\uparrow$(\%) & 12.27& \textcolor{red}{\xmark} & \textcolor{red}{\xmark} & \textbf{78.13}\\
        & PE$\downarrow$(cm) & 2.81& \textcolor{red}{\xmark} & \textcolor{red}{\xmark} & \textbf{0.53}\\
        & OE$\downarrow$($^\circ$) & 28.77& \textcolor{red}{\xmark} & \textcolor{red}{\xmark} & \textbf{18.74}\\
        & AS$\uparrow$ & 4.24& \textcolor{red}{\xmark} & \textcolor{red}{\xmark} & \textbf{8.93}\\
        \midrule
        \multirow{4}{*}{\makecell{3 (1 \\ hand)}} 
        & SR$\uparrow$(\%) & 4.36& \textcolor{red}{\xmark} & \textcolor{red}{\xmark} & \textbf{71.81}\\
        & PE$\downarrow$(cm) & 2.65& \textcolor{red}{\xmark} & \textcolor{red}{\xmark} & \textbf{0.89}\\
        & OE$\downarrow$($^\circ$) & 29.19& \textcolor{red}{\xmark} & \textcolor{red}{\xmark} & \textbf{12.95}\\
        & AS$\uparrow$ & 2.48& \textcolor{red}{\xmark} & \textcolor{red}{\xmark} & \textbf{9.13}\\
        \midrule
        \multirow{4}{*}{\makecell{5 (1 \\ hand)}} 
        & SR$\uparrow$(\%) & 5.68& 26.44& \textcolor{red}{\xmark} & \textbf{85.67}\\
        & PE$\downarrow$(cm) & 2.89& 1.93& \textcolor{red}{\xmark} & \textbf{0.77}\\
        & OE$\downarrow$($^\circ$) & 29.13& 22.28& \textcolor{red}{\xmark} & \textbf{10.90}\\
        & AS$\uparrow$ & 3.07& 5.88& \textcolor{red}{\xmark} & \textbf{9.29}\\
        \midrule
        \multirow{4}{*}{\makecell{5 (2 \\ hands)}} 
        & SR$\uparrow$(\%) & 2.10& 28.75& 25.33& \textbf{57.67}\\
        & PE$\downarrow$(cm) & 2.96& 1.44& 1.66& \textbf{0.72}\\
        & OE$\downarrow$($^\circ$) & 29.56& 16.84& 14.13& \textbf{13.44}\\
        & AS$\uparrow$ & 2.58& 4.34& 4.81& \textbf{8.16}\\
        \bottomrule
        \end{tabular}
    }

\end{table}

\noindent \textbf{Experimental equipment.} All experiments were conducted on servers equipped with an NVIDIA GeForce RTX 4090 GPU and an Intel Core i9‑14900K CPU. We evaluated our methods on the Franka 2‑fingered gripper, the xArm 2‑fingered gripper, the CASIA Hand‑G 3‑fingered dexterous hand\cite{yan2025casiahand}, the Inspire 5‑fingered dexterous hand, and the OHand$^{TM}$ dexterous hand.

\subsection{Comparison with SOTA}
\label{sec:com}
To comprehensively evaluate UniBYD, we compare it with three representative baselines: a classical retargeting method based on optimization-based inverse kinematics, and two current SOTA methods for dexterous manipulation, ManipTrans \cite{li2025maniptrans} and DexMachina* \cite{mandi2025dexmachina}. Since existing SOTA methods do not support 2/3-fingered tasks, and there is a lack of comparative approaches for learning 2/3-fingered dexterous manipulation from human demonstrations, we employ retargeting as the primary baseline for such embodiments. 
To ensure a fair comparison, we reimplement DexMachina within our simulator, denoted as \textit{DexMachina}$^*$ throughout this paper. To maintain consistency, all methods share the same experimental configuration, the implementation details of which are provided in Appendix Section 9. 
The comparative results are reported in \cref{tab:1}, where \textcolor{red}{\xmark} denotes that the method does not support that hand type.

The results unequivocally demonstrate that UniBYD surpasses all baselines across every evaluation dimension, achieving a 44.08\% improvement in success rate over ManipTrans (SOTA). This underscores UniBYD's superior high-dimensional coordination, whereas imitation-centric ManipTrans is constrained by the embodiment gap and goal-centric DexMachina struggles with complex collaboration without fine-grained guidance. First, UniBYD is the only unified framework that consistently achieves high reliability across all hand morphologies, whereas ManipTrans and DexMachina do not support 2/3-fingered hands. 
Second, retargeting yields marginal success rates. 
Even on challenging 5-finger bimanual tasks, UniBYD achieves a 57.67\% success rate, outperforming the best baseline by a substantial 28.92\% margin. 
On 5-fingered unimanual tasks, UniBYD achieves an 85.67\% SR, outperforming ManipTrans by a 59.23\% margin. Simultaneously, UniBYD significantly improves manipulation precision, reducing PE and OE by 60.10\% and 51.08\% compared to ManipTrans, demonstrating its superior fine-grained control.
On 2/3-fingered tasks, UniBYD achieves a high SR of 78.13\% and 71.81\%, respectively. This underscores its robust cross-embodiment generalization and ability to complete tasks reliably.
Finally, UniBYD consistently maintains a superior AS ($\ge$ 8.16), markedly exceeding all baselines (max 5.88). This proves that its learned policies are not only more effective but also more strongly aligned with specific embodiments. 

\begin{table}[t]
    \small
    \centering
    \caption{Ablation study of UniBYD. The results validate the individual contributions of SE and GR, as well as the incremental gain of LSC, in enhancing performance.}
    \label{tab:2}
    \resizebox{\linewidth}{!}{
        \setlength{\tabcolsep}{2pt} 
        \begin{tabular}{@{}llccccc@{}} 
        \toprule
        \makecell{\textbf{Hand} \\ \textbf{Type}} &
        \textbf{Metrics} &
        \textbf{base} &
         \textbf{+SE}&
        \textbf{+GR} &
        \makecell{\textbf{+GR} \\ \textbf{+LSC}}  &\textbf{UniBYD}\\ 
        \midrule
        \multirow{4}{*}{\makecell{2 (1 \\ hand)}} 
        & SR$\uparrow$(\%) & 24.94& 67.06& 56.19& 61.50 &\textbf{78.13}\\
        & PE$\downarrow$(cm) & 2.66& 0.54& 1.28& 1.32 &\textbf{0.53}\\
        & OE$\downarrow$($^\circ$) & 27.51& 19.89& 21.31& 21.58 &\textbf{18.74}\\
        & AS$\uparrow$ & 5.63& 7.56& 7.99& 8.36&\textbf{8.93}\\
        \midrule
        \multirow{4}{*}{\makecell{3 (1 \\ hand)}} 
        & SR$\uparrow$(\%) & 19.56& 51.06& 65.13& 67.63&\textbf{71.81}\\
        & PE$\downarrow$(cm) & 2.45& 0.95& 0.93& \textbf{0.87}&0.89\\
        & OE$\downarrow$($^\circ$) & 26.09& 14.67& 14.06& \textbf{12.85}&12.95\\
        & AS$\uparrow$ & 5.42& 6.87& 8.52& 8.78&\textbf{9.13}\\
        \midrule
        \multirow{4}{*}{\makecell{5 (1 \\ hand)}} 
        & SR$\uparrow$(\%) & 13.56& 55.11& 63.17& 65.39&\textbf{85.67}\\
        & PE$\downarrow$(cm) & 2.46& 1.24& 0.87& 0.79&\textbf{0.77}\\
        & OE$\downarrow$($^\circ$) & 27.12& 17.29& 13.22& 13.23&\textbf{10.90}\\
        & AS$\uparrow$ & 4.87& 6.33& 8.29& 8.64&\textbf{9.29}\\
        \midrule
        \multirow{4}{*}{\makecell{5 (2 \\ hands)}} 
        & SR$\uparrow$(\%) & 33.46& 43.33& 47.01& 54.17&\textbf{57.67}\\
        & PE$\downarrow$(cm) & 1.46& 1.43& 1.48& 1.36&\textbf{0.72}\\
        & OE$\downarrow$($^\circ$) & 15.35& 17.64& 17.17& 14.17&\textbf{13.44}\\
        & AS$\uparrow$ & 3.20 & 3.55 & 6.27 & 6.73&\textbf{8.16}\\
        \bottomrule
        \end{tabular}
    }
\end{table}

\begin{figure}[t] 
    \centering
    \includegraphics[width=\linewidth]{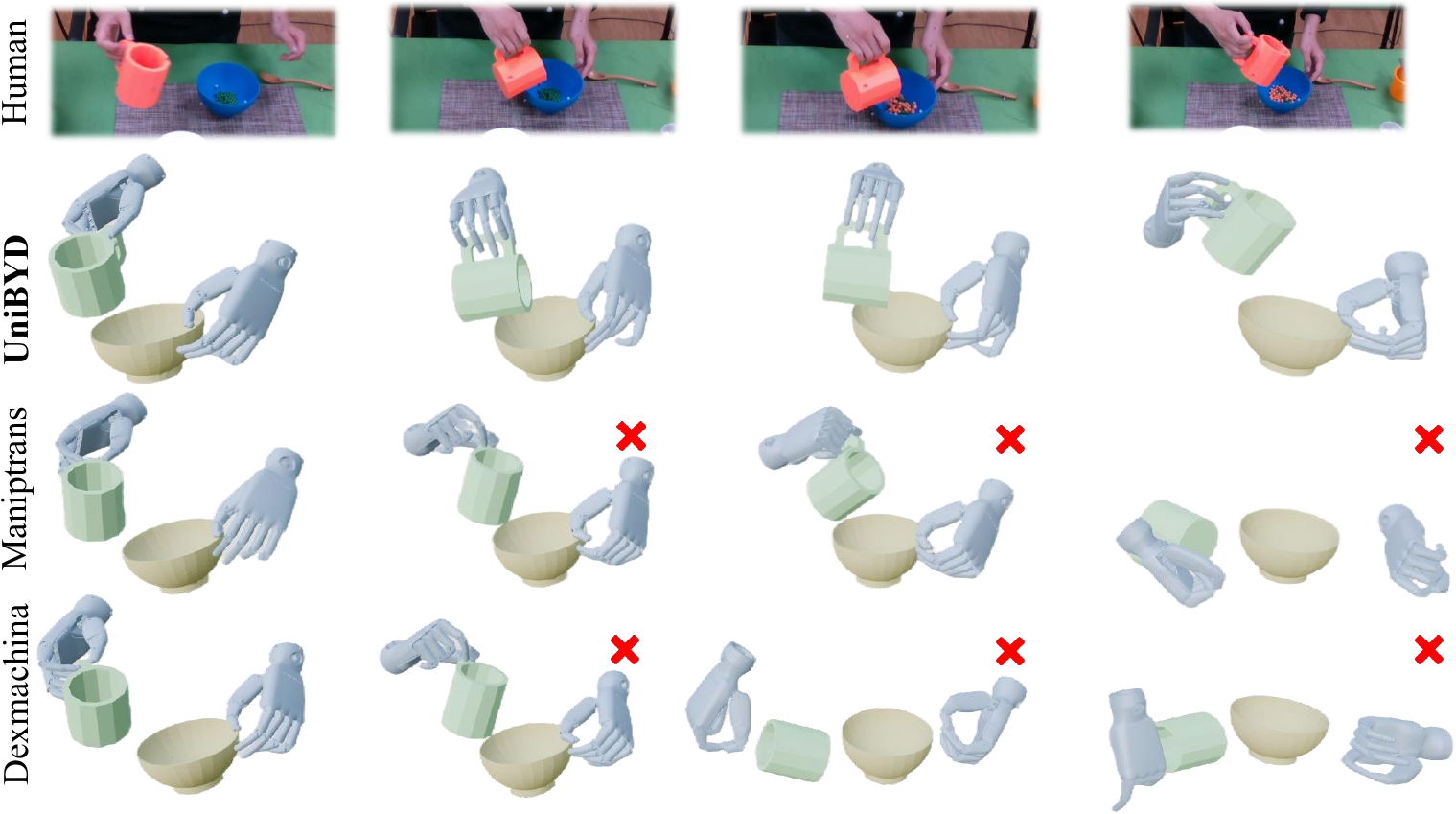}
    \caption{A visual comparison of the experimental results. UniBYD learns manipulation strategies aligned with the robot's embodiment, thereby successfully completing the task, whereas both ManipTrans and DexMachina* fail.}
    \label{fig:com}
\end{figure}

\begin{figure}[t] 
    \centering
    \includegraphics[width=\linewidth]{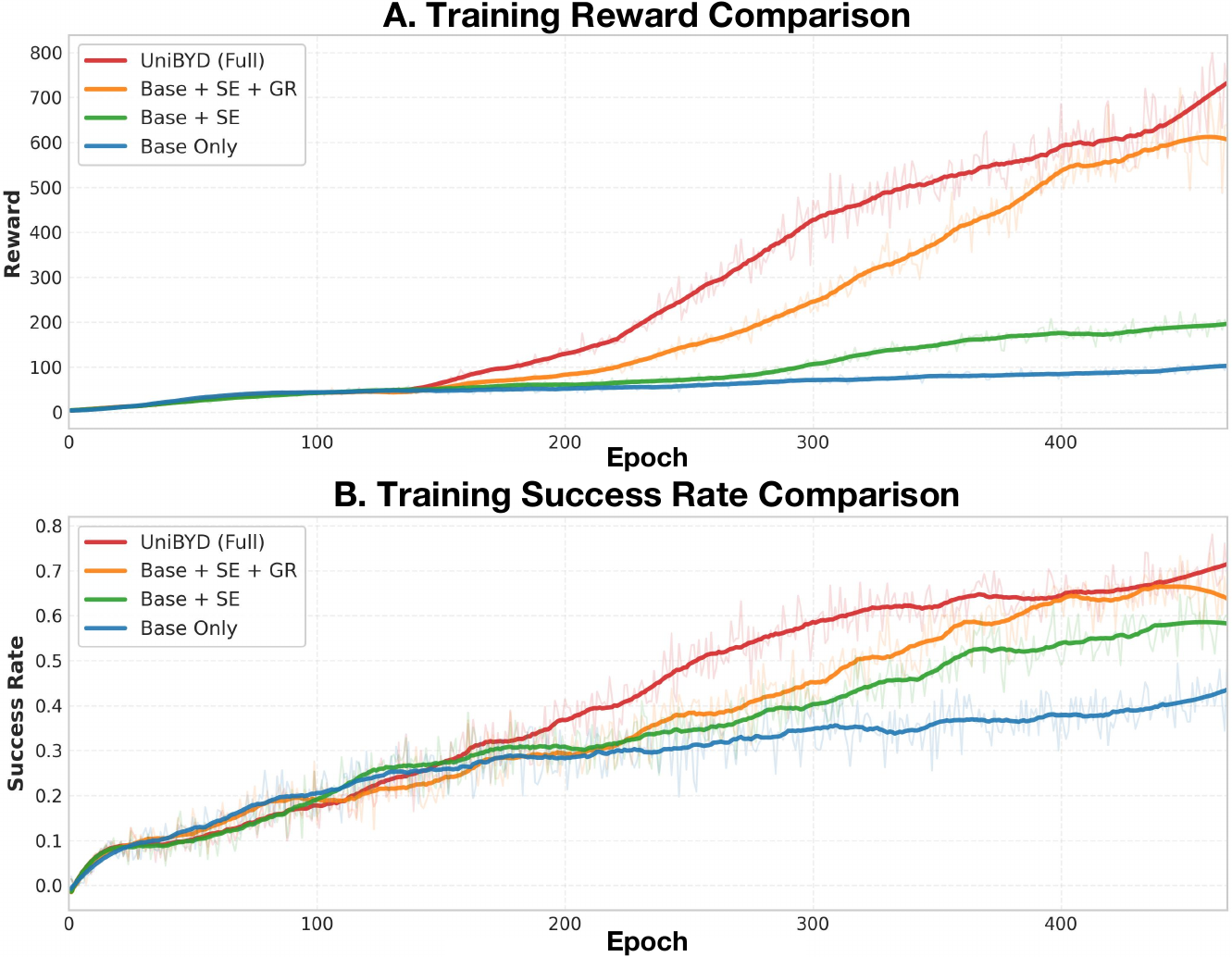}
    \caption{Training procedure of a representative task.}
    \label{fig:xiaorongquxian}
\end{figure}

As illustrated in \cref{fig:com}, we visually compare the manipulation strategies learned by UniBYD against those of other methods. ManipTrans seeks to emulate the human tactic of grasping the mug with the three rear fingers. However, these fingers are too wide to pass through the handle, causing the mug to slip and fall. Hampered by sparse reward signals, DexMachina* likewise fails to discover a viable strategy. By contrast, through trial and error, UniBYD identifies that the three robotic fingers are much wider than those of a human. It adapts by using only two fingers (the middle and ring fingers) to grasp the handle, pinching it with the thumb and index finger while bracing the mug with the little finger.

\subsection{Ablation Study}
As reported in \cref{tab:2}, we conduct an ablation study built upon the base model. We define the base as an RL approach utilizing only the imitation rewards, functionally equivalent to behavior cloning and serving as the primary baseline for \cref{sec:com}. Notably, even base outperforms existing SOTA in bimanual tasks, confirming that our UMR and imitation rewards provide a superior foundation.
Compared to the base, +SE (base with \textit{shadow engine}) improves the success rate by 31.26\%, validating that the fine-grained expert guidance provided by SE is essential for effective policy initialization.
+GR (base with goal reward $R^{goal}$) achieves a 35.00\% SR gain and elevates AS to 7.77. This variant facilitates a seamless transition from offline-data-driven imitation to online goal-driven exploration. This allows the robot to transcend human priors and discover physically optimal postures uniquely tailored to its own morphology.
+GR+LSC (base with $R^{goal}$ and loss synergy with counterbalancing) maintains strategic stochasticity, effectively preventing premature convergence to suboptimal imitation. More investigations into other components, including manipulator-specific information(UMR), hyperparameter sensitivity, Boundary Loss, Entropy Loss, and the Reward Annealing schedule are provided in Appendix Section 14.

\cref{fig:xiaorongquxian} illustrates the evolution of reward and success rate in a representative task. Base is unable to explore optimal postures, rapidly settling into a suboptimal policy. In contrast, +SE more accurately reproduces human behavior. Building upon +SE, the introduction of $R^{goal}$ forms +SE+GR. With the fading of expert guidance and under the direction of explicit objectives, this strategy explores postures that lead to higher success rates. Furthermore, combined with LSC (UniBYD), it encourages exploration by maintaining stochasticity in the early phase, eventually converging to a superior manipulation strategy.
\begin{figure*}[t]
  \centering
  \includegraphics[width=\linewidth]{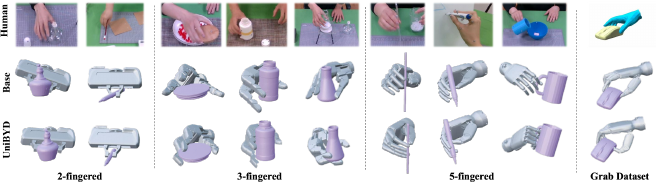}
   \caption{Comparative experimental results for base and UniBYD.}
   \label{fig:chaoyue}
\end{figure*}

\subsection{Qualitative Analysis: Beyond Mere Imitation}

As shown in \cref{fig:chaoyue}, UniBYD can discover manipulation strategies aligned with each robot’s embodiment.
For the 2-fingered gripper, the imitation-only base rigidly reproduces the human’s oblique grasp, while UniBYD adopts a more stable grasp oriented perpendicular to the body of the alcohol burner.
For the 3-fingered hand, unlike the base that relies solely on two fingertips to pinch the Erlenmeyer flask, UniBYD secures the flask with two fingers while employing the third finger to support the bottom for a more stable manipulation.
For the 5-fingered hand, in tasks such as stirring and handwriting, the base imitates a two-fingertip pinch prone to slippage, whereas UniBYD employs the remaining three digits to provide supportive contact forces, enhancing success.
To demonstrate the generalizability of UniBYD, we further present results on the GRAB dataset \cite{taheri2020grab} in \cref{fig:chaoyue}, where UniBYD achieves a remarkable 98.00\% SR and 9.43 AS.

\begin{figure}[t] 
    \centering
    \begin{minipage}{0.48\textwidth}
        \centering
        \includegraphics[width=\linewidth]{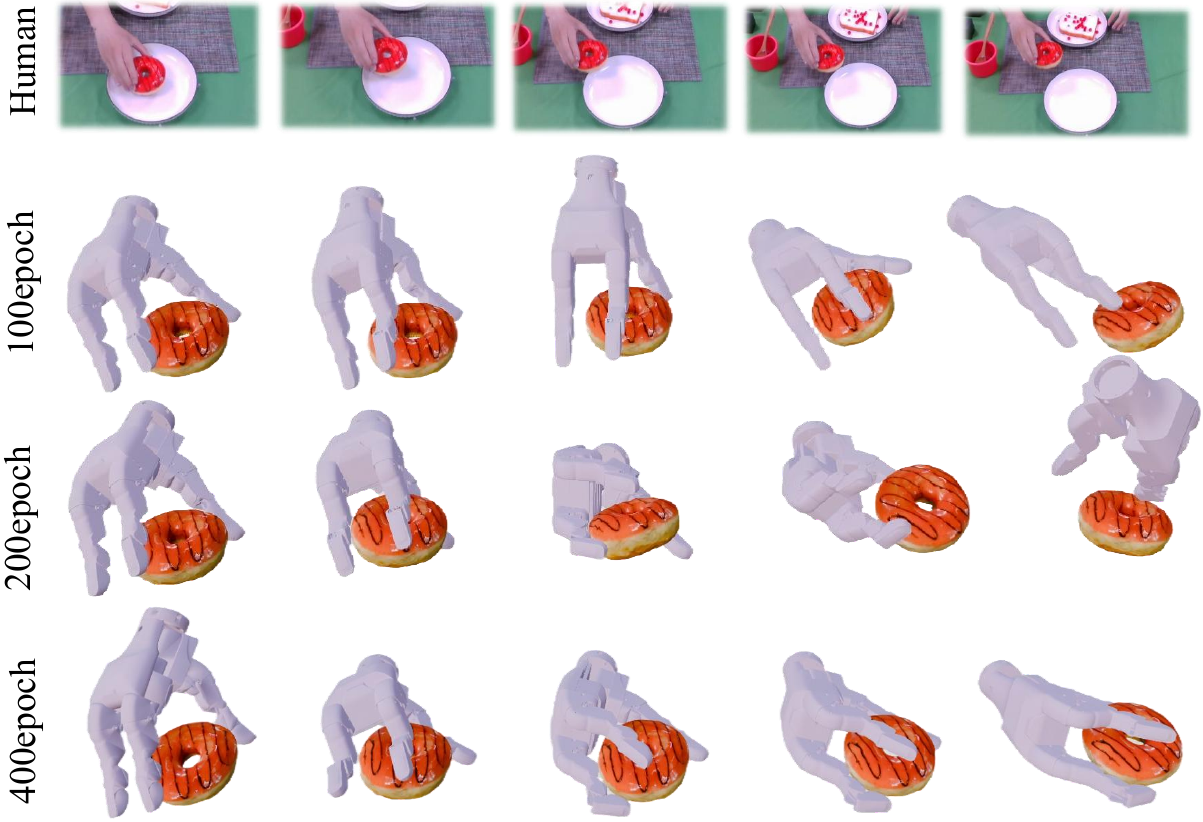}
        \caption{Results illustrating the progressive evolution of manipulation strategies over the course of training.}
        \label{fig:epoch}
    \end{minipage}
    \hfill 
    \begin{minipage}{0.48\textwidth}
        \centering
        \includegraphics[width=\linewidth]{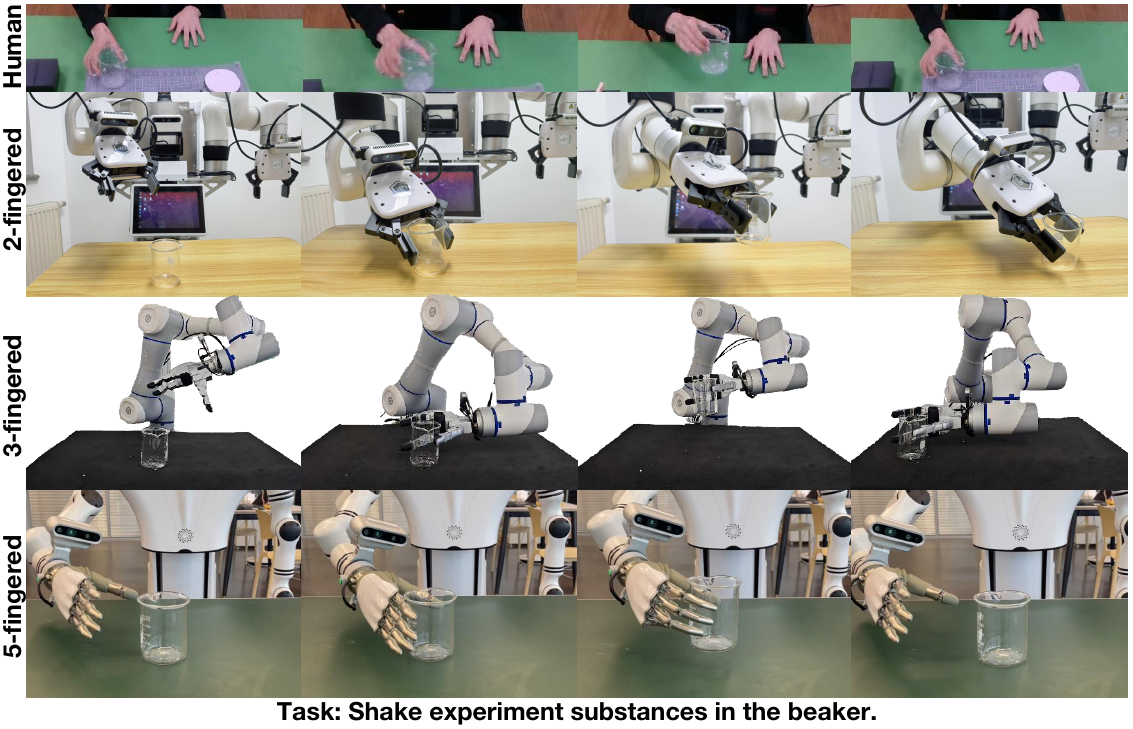}
        \caption{Experimental results on real-world robotic platforms.}
        \label{fig:final}
    \end{minipage}
\end{figure}

To investigate the progression from imitation to exploration, we compare checkpoints across training epochs, as shown in \cref{fig:epoch}. By epoch 100, training is dominated by imitation. The policy attempts to manipulate with two fingers but frequently drops it due to limited force control. By epoch 200, the policy begins to relax its reliance on expert data, attempting to grasp the doughnut with the first and third digits while probing the role of the second digit. By epoch 400, it discovers a strategy that pinches the doughnut with two fingers and braces it with the remaining finger. Exploiting the mechanical characteristics, UniBYD observes that operating with the wrist canted sideways enhances success.

\subsection{Real-World Experiments}

By mapping the dexterous hand’s wrist to the robotic arm flange and aligning its degrees of freedom, we conduct experiments on three real-world platforms: the X-Arm 2-fingered hand, the Casia Hand-G 3-fingered dexterous hand, and the OHand$^{TM}$ 5-fingered dexterous hand. To facilitate zero-shot transfer, we employ FoundationPose \cite{wen2024foundationpose} for hand-object alignment and MoveIt! for trajectory planning. More implementation details are provided in Appendix Section 9.5.

The system achieves success in 26 of 50 trials, 32 of 50 trials, and 35 of 50 trials, respectively. As shown in \cref{fig:final}, for the same task, UniBYD tailors its manipulation strategy to the end-effector morphology. With a 2-fingered gripper, it clamps the beaker diagonally, whereas with a 3-fingered hand, it wraps the beaker with all fingers. These results prove that UniBYD transfers effectively to the real world, despite a slight drop in success rate. Failure analysis indicates that failure modes include hand-object collisions (19\%), object drops (9\%), joint limit triggers (7\%), and hand-table collisions (3\%).

\section{Conclusion}
We introduce UniBYD, a unified framework that learns robotic manipulation beyond imitation from human demonstrations. 
UniBYD achieves a systematic evolution from fine-grained imitation to morphology-adaptive exploration, discovering manipulation strategies that align with each robot’s embodiment.
Real-world results confirm its superior embodiment alignment and transferability.

{
    \small
    \bibliographystyle{ieeenat_fullname}
    \bibliography{main}
}
\clearpage
\setcounter{page}{1}
\maketitlesupplementary

This supplementary material provides comprehensive implementation details and additional experimental results. 
We begin by elaborating on the core mechanisms of the Hybrid Markov-based \textit{shadow engine} and the associated dynamic control strategies in \cref{sec:se}. 
\cref{sec:cir} and \cref{sec:sfc} provide the precise mathematical definitions for the comprehensive reward functions and the curriculum-based success/failure criteria, respectively. 
\cref{sec:ed} outlines the specific experimental hyperparameters, the optimization-based motion retargeting pipeline, and the reproduction details for the baseline methods (ManipTrans and DexMachina). 
\cref{sec:uni} details the MLLM-driven pipeline for generating cross-morphology expert data and the evaluation metrics for the UniManip benchmark. 
\cref{sec:pe} analyzes the evolution of success rate and episode length over the training course to validate the effectiveness of UniBYD. 
\cref{sec:al} presents the complete pseudocode of the UniBYD framework. 
\cref{sec:dis} demonstrates the distinct manipulation strategies learned by UniBYD tailored to different robotic morphologies for the same task. 
Finally, \cref{sec:me} provides extensive additional qualitative results across diverse robotic hands in both simulation and the real world.

\section{Hybrid Markov-based Shadow Engine}
\label{sec:se}
\subsection{Dexterous Hand Control}
\label{sec:sehand}
The standard training process functions as a pure Markov process where the state $s_{t+1}$ at step $t+1$ is fully determined by the action $\Delta a_t^{\pi}$. Specifically, $\Delta a_t^{\pi}$ is predicted by the policy $\pi_{\theta}$ based on the observation $o_t$ at step $t$, denoted as $\Delta a_t^{\pi} = \pi_{\theta}(o_t)$. Under this training paradigm, state updates are driven exclusively by the actions predicted by the model itself. However, during the early stages of training, the model possesses limited capabilities, resulting in significant errors in the predicted actions at each step. The compounding errors accumulated over just a few consecutive steps are sufficient to cause the object to deviate completely from the target state, leading the episode to be immediately judged as a failure and terminated. Such frequent episode terminations and resets severely degrade the efficiency and capacity of the model for early knowledge acquisition. This issue is particularly critical for complex manipulation tasks. If the model fails to grasp preliminary manipulation skills early on, combined with the fact that success criteria become increasingly stringent as training epochs increase, failures in the mid-to-late stages will become more frequent. Consequently, the number of continuous steps learned within an episode decreases, rendering it impossible for the model to learn the complete execution flow of the task.

Therefore, to ensure that the model can preliminarily grasp manipulation skills during the early stages and to enhance training efficiency, we propose a Hybrid MDP mechanism. In the first training epoch, where $\beta_t = 1$ and $\alpha_t = 0$, UniBYD still uses the predicted action $\Delta a_t^{\pi}$ at step $t$ to compute the reward function. However, it exclusively employs the expert demonstration action $\Delta a_t^{E}$ to update the simulator and obtain the next state. Effectively, the model learns the manipulation at each individual step in isolation, unaffected by the previous step. This phase approximates discrete pointwise learning, focusing solely on the action at each step while the state update relies entirely on expert data.

In the early hybrid phase, exemplified by $\beta_t = 0.7$ and $\alpha_t = 0.3$, the action $\Delta a_t^{exec}$ executed by the simulator is a weighted blend of the policy-predicted action and the expert action, defined as $\Delta a_t^{exec} = 0.3 \cdot \Delta a_t^{\pi} + 0.7 \cdot \Delta a_t^{E}$. This implies that the evolution of the subsequent state $s_{t+1}$ begins to be partially influenced by the model's own policy $\pi_{\theta}$. Consequently, the state distribution of the environment is no longer identical to the expert demonstration trajectory but starts to incorporate perturbations introduced by the policy itself. However, as the weight of the expert action $\beta_t$ remains dominant, action errors resulting from model predictions are significantly corrected by the \textit{shadow engine}. While allowing the model to explore action-state transition logic, this mechanism employs strong expert guidance to constrain the states of the robotic hand and the object within a safe region. This effectively prevents error accumulation and catastrophic state deviations caused by the immaturity of the policy in the early stages. This establishes a controlled, error-tolerant learning environment that enables the model to sustain episode continuity in long-horizon tasks, thereby allowing it to observe states in the later stages of the process.

In the late hybrid stage, exemplified by $\beta_t = 0.3$ and $\alpha_t = 0.7$, the dominance of action control shifts. At this point, the executed action $\Delta a_t^{exec}$ is primarily determined by the prediction $\Delta a_t^{\pi}$ from the policy network $\pi_{\theta}$, with the expert demonstration action $\Delta a_t^{E}$ serving merely as an auxiliary correction term. This implies that the update of the environmental state $s_{t+1}$ depends largely on the model's own decisions, and the state distribution closely approximates the true distribution observed under fully autonomous mode. During this phase, the expert action provides only a weak "corrective" signal and can no longer fully mask the accumulated errors generated by the model. This forces the policy network to learn to handle and rectify state deviations induced by its own actions, thereby maintaining manipulation stability within a nearly authentic Markov chain. This design ensures that the model adapts to self-induced distribution shift, effectively preparing it for the final transition where $\beta_t$ drops to 0, the assistance of the \textit{shadow engine} is completely withdrawn, and the system enters a pure Markov process.

As training epochs increase, we adhere to a linear decay schedule defined as $\beta_t = \max\left(0, 1 - \frac{e}{T_{decay}}\right)$, gradually reducing the weight of expert guidance. When $\beta_t$ eventually decays to 0 and $\alpha_t$ rises to 1, the \textit{shadow engine} completely disengages, allowing the model to transition smoothly into a full Markov Decision Process. At this stage, the model has acquired preliminary competencies and must independently shoulder the long-horizon cumulative consequences resulting from all actions, thereby completing a robust transition from discrete pointwise imitation to continuous autonomous decision-making.

Through this smooth transition from discrete pointwise learning to hybrid guided learning, and finally to fully autonomous sequential learning, UniBYD effectively addresses the exploration challenges in dexterous manipulation tasks caused by the initially weak policy in reinforcement learning.

\subsection{Object Control}
\label{sec:seobject}
\begin{figure}[t]
  \centering
  \includegraphics[width=0.7\linewidth]{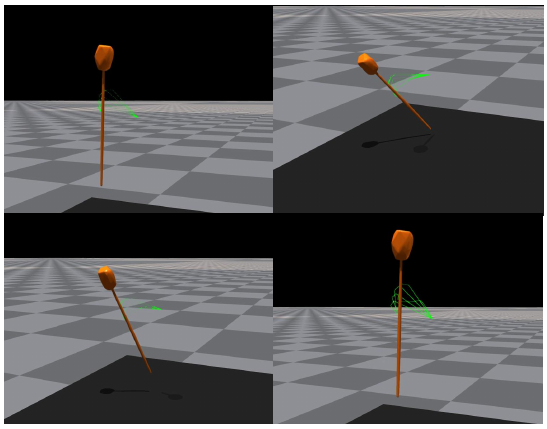}

   \caption{An example of task completion using a PD controller without robotic hands. The green fingertip keypoints shown in the figure represent the location of the robotic hand in the expert demonstration data.}
   \label{fig:pd}
\end{figure}

To mitigate the instability of object manipulation during the early training phase and prevent premature episode termination caused by object drops, the \textit{shadow engine} applies a dynamic auxiliary force $F_{support}$ to the object. As shown in \cref{fig:pd}, when a substantial support force is applied to the object, the object can be constrained to move near its target trajectory, preventing it from falling.

At each time step $t$, we first retrieve the target object pose $g_t^{obj}$ and velocity $\dot{g}_t^{obj}$ from the expert demonstration, which serve as the reference trajectory. We then compute the supporting force using a PD controller based on the deviation of the current object state from this reference:
\begin{equation}
F_{support} = K_{p,e} (g_t^{obj} - o_t^{obj}) + K_{d,e} (\dot{g}_t^{obj} - \dot{o}_t^{obj}),
\label{eq:important}
\end{equation}
where $o_t^{obj}$ and $\dot{o}_t^{obj}$ denote the observed object pose and velocity at time $t$, respectively. $K_{p,e}$ and $K_{d,e}$ are the proportional and derivative gain coefficients at the current training epoch $e$.

Considering the physical differences between various manipulated objects, we normalize the PD parameters according to the actual object mass $m_{obj}$. Given a reference object mass $m_{ref}$ and predefined baseline gains $K_{p,cfg}$ and $K_{d,cfg}$, the initial PD parameters $K_p^{start}$ and $K_d^{start}$ are calculated as follows:
\begin{equation}
K_p^{start} = K_{p,cfg} \cdot \frac{m_{obj}}{m_{ref}}, \quad K_d^{start} = K_{d,cfg} \cdot \frac{m_{obj}}{m_{ref}}.
\label{eq:pd_init}
\end{equation}

Crucially, to ensure that the object dynamics transition to a purely physical interaction state synchronously with the policy assuming full control authority, the gain coefficients follow the same linear decay schedule as the hand action blending weight $\beta_t$. The decay is defined as:
\begin{equation}
K_{p,e} = K_p^{start} \cdot \max\left(0, 1 - \frac{e}{T_{decay}}\right)
\label{eq:important},
\end{equation}
\begin{equation}
K_{d,e} = K_d^{start} \cdot \max\left(0, 1 - \frac{e}{T_{decay}}\right),
\label{eq:important}
\end{equation}
where $T_{decay}$ is the predefined decay horizon, which is identical to the threshold used in the Dexterous Hand Control module. This synchronization guarantees that the auxiliary support force $F_{support}$ is completely removed ($K_{p,e}=0, K_{d,e}=0$) exactly when the \textit{shadow engine} ceases its guidance ($e \ge T_{decay}$), forcing the policy to maintain object stability independently.

\section{Computation of the Imitation Reward}
\label{sec:cir}
The imitation reward $R_t^{imitation}$ quantifies the similarity between the current state and the expert demonstration. This reward function comprises a weighted sum of state-similarity components and a power penalty term for action regularization. For each reward component $r_k$ in the summation, we employ a standardized exponential kernel defined generally as $r_k(s_t, s_t^E) = \exp(-\lambda_k \cdot E_k)$. Here, $E_k$ represents the specific physical error metric, and $\lambda_k$ denotes the sensitivity coefficient, which governs the decay rate of the reward with respect to the error.

We first impose strict constraints on the kinematic state of the robotic hand's wrist. Specifically, the wrist position component $r_{eef\_pos}$ measures the Euclidean distance between the current wrist position $p_{eef}$ and the expert target position $p_{eef}^{E}$. To ensure tracking precision, this component is assigned a weight $w_{eef\_pos}$ of $0.1$ and a high sensitivity coefficient $\lambda_{eef\_pos}$ of $40$, with the formula defined as $r_{eef\_pos} = \exp(-40 \cdot \|p_{eef} - p_{eef}^{E}\|_2)$. Simultaneously, the wrist orientation component $r_{eef\_rot}$ utilizes quaternion differences to quantify the rotational deviation $\theta_{diff}$. Given the critical role of orientation in manipulation stability, this component is assigned a higher weight of $0.6$ and a sensitivity coefficient of $1$. The calculation is $r_{eef\_rot} = \exp(-1 \cdot \theta_{diff})$, where the deviation angle $\theta_{diff}$ is the magnitude of the axis-angle representation extracted from the relative rotation quaternion.

To further ensure motion smoothness and consistency, we incorporate velocity synchronization rewards, calculated based on the mean absolute error across dimensions. The end-effector linear velocity component $r_{eef\_vel}$ computes the average deviation of the Cartesian velocity vector, with a weight of $0.1$ and a sensitivity coefficient of $1$. Similarly, the end-effector angular velocity component $r_{eef\_ang\_vel}$ applies the same constraint to the 3D angular velocity vector but with a reduced weight of $0.05$ to balance the optimization objectives. Finally, to regularize the internal joint movements, the joint velocity component is defined as $r_{joints\_vel} = \exp(-0.1 \cdot \frac{1}{N}\sum_{j=1}^N |\dot{q}_j - \dot{q}_j^{E}|)$. This term is normalized with a weight of $0.1$ and a sensitivity coefficient of $0.1$, where $N$ represents the total number of degrees of freedom, and $\dot{q}_j$ and $\dot{q}_j^{E}$ denote the current and expert angular velocities of the $j$-th joint, respectively. This formula aims to minimize the average velocity deviation across the entire joint space.

Addressing the crucial aspect of fingertip control in dexterous manipulation, we implement a fine-grained hierarchical weighting scheme. Each fingertip position reward $r_{tip}$ is calculated based on the Euclidean distance between the current fingertip position $p_{tip}$ and the expert target $p_{tip}^{E}$ using the general form $r_{tip} = \exp(-\lambda_{tip} \cdot \|p_{tip} - p_{tip}^{E}\|_2)$. As the primary drivers for stable opposition and pinching, the thumb and index finger are accorded the highest priority. Specifically, the thumb component $r_{thumb}$ is assigned a weight of $0.9$ with a high sensitivity coefficient $\lambda_{thumb}$ of $100$ to ensure extreme positional precision. The index finger component $r_{index}$ follows with a weight of $0.8$ and a sensitivity coefficient of $90$. The middle finger, serving as a primary support, is assigned a weight of $0.75$ and a sensitivity of $80$, while the ring and little fingers, acting as auxiliary supports, are each assigned a weight of $0.6$ and a sensitivity of $60$. Furthermore, to encourage necessary physical interaction, we include a fingertip contact force component $r_{force}$ with a weight of $1.0$. This component is defined as $r_{force} = \exp(-1 \cdot (\|F_{tips}\|_2 + \epsilon)^{-1})$, where $F_{tips}$ represents the resultant vector of effective contact forces on all fingertips and $\epsilon$ is a stability constant. This inverse exponential mapping penalizes excessive impact forces while imposing a severe penalty when contact force approaches zero, thereby enforcing sustained stable contact with the object.

Regarding the manipulated object, we guide the policy to reproduce the expert's trajectory through multi-dimensional state constraints. The object position component $r_{obj\_pos}$ is calculated based on the Euclidean distance between the current object position $p_{obj}$ and the expert target $p_{obj}^{E}$. To ensure millimeter-level tracking precision, this component is assigned the highest weight in the entire reward system ($5.0$), combined with a high sensitivity coefficient of $80$, defined as $r_{obj\_pos} = \exp(-80 \cdot \|p_{obj} - p_{obj}^{E}\|_2)$. Simultaneously, the object orientation component $r_{obj\_rot}$ aims to minimize the rotational deviation between the current pose quaternion $q_{obj}$ and the target quaternion $q_{obj}^{E}$. This component is assigned a weight of $1.0$ and a sensitivity coefficient of $3$. Additionally, to ensure the smoothness of object motion, we introduce velocity synchronization rewards. The object linear velocity component $r_{obj\_vel}$ and angular velocity component $r_{obj\_ang\_vel}$ calculate the mean absolute deviations of the velocity vectors in 3D space. Both are assigned a weight of $0.1$ and a sensitivity coefficient of $1$, constraining the dynamic characteristics of the object during manipulation.

Finally, regarding the penalty term $p(\Delta a_t)$, we define it as a regularization constraint on the system's energy consumption. Specifically, it calculates the sum of the absolute products of joint torques $\tau$ and angular velocities $\dot{q}$, given by $P = \sum |\tau \cdot \dot{q}|$. To strike a balance between minimizing energy expenditure and successful task completion, this term is scaled by a weight coefficient of $0.5$. This design is intended to prevent the policy from generating high-frequency oscillations or unrealistic violent movements, thereby guiding the model to learn manipulation strategies that are both natural and energy-efficient.

\section{Success and Failure Criteria}
\label{sec:sfc}
The sparse goal reward $R^{goal}$ is activated exclusively when the episodic success indicator $\mathcal{C}_{suc}=1$. To ensure policy robustness and preclude accidental successes, we define $\mathcal{C}_{suc}$ as a dual criterion function grounded in trajectory completion and dynamic state constraints. Specifically, a training episode is deemed successful if and only if the current execution step $t$ approaches the total length of the expert demonstration $T_{max}$, and no failure conditions are triggered within a stability window $T_{stable}$ (default set to 3 steps) at the trajectory's conclusion:
\begin{equation}
\mathcal{C}_{suc} = \mathbb{I}\left((t + 1 + T_{stable} \ge T_{max}) \land (\neg \text{Failed}) \right).
\label{eq:important}
\end{equation}

The failure determination logic, denoted as $\text{Failed}$, is not static but is tightly coupled with the training phase divisions described in the main text. During the early imitation-driven reinforcement learning phase, the policy has not yet acquired basic manipulation competence. When the average success rate $\bar{SR}$ is below the predefined threshold, we impose high-dimensional geometric constraints. These constraints guide the policy to accurately reproduce the expert's hand configuration. In this phase, a failure is triggered if the current step exceeds the minimum step count $t_{min}$ (designed to ignore initial perturbations) and any of the following conditions are met. Firstly, the object position deviates severely from the target, meaning the Euclidean distance between the current object position $p_{obj}$ and the expert target $p_{obj}^E$ exceeds a dynamic threshold: $\|p_{obj} - p_{obj}^E\|_2 > \delta_{pos} \cdot S^3$, where $\delta_{pos}$ is the base position threshold and $S$ is the dynamic scaling factor. Secondly, the object orientation deviation is excessive, such that the axis-angle magnitude $\theta_{diff}$ derived from the quaternion difference satisfies $\theta_{diff} > \delta_{rot} \cdot S^3$. Thirdly, to ensure the correct grasping configuration, the fingertip positions of key digits (thumb, index, and middle fingers) are required to precisely track the expert trajectory: $\|p_{tip} - p_{tip}^E\|_2 > \delta_{finger} \cdot S$. Finally, we detect unintended collisions, defined as instances where the distance $d_{tip}$ between any fingertip and the object is less than the collision threshold $\epsilon_{coll}$ while no contact exists in the expert demonstration ($\neg c_{target}$): $(d_{tip} < \epsilon_{coll}) \land (\neg c_{target})$.

As the model's capabilities improve, the training process advances into the hybrid phase and ultimately the exploration phase. During these stages, constraints imposed by expert data are gradually relaxed to encourage the policy to discover solutions adapted to the robot's specific morphology. Once the success rate $\bar{SR}$ exceeds the threshold, the failure criteria no longer enforce strict constraints on fingertip positions and unintended collisions. Instead, the focus shifts to the core task objectives and the physical stability of the manipulation. At this point, while retaining the deviation constraints on object position and pose mentioned above, we introduce a constraint on the number of contact points. A failure is declared when the number of effective contact points on the object's surface, denoted as $N_{contact}$, falls below the minimum required for a stable grasp, $N_{min}$ (specifically, 2 for unimanual tasks and 3 for bimanual tasks): $N_{contact} < N_{min}$.

This evolution in the determination logic ensures a smooth transition of the policy from early rote imitation to late-stage goal-oriented autonomous exploration. The term $S$ in the aforementioned formulas represents the dynamic scale factor. It functions as a time-dependent variable that decays over the course of training, thereby implementing an automatic curriculum learning mechanism. Its computation adheres to an exponential decay schedule:
\begin{equation}
S(t) = (e \cdot 2)^{-t/T_{tighten}} \cdot (1 - S_{final}) + S_{final},
\label{eq:important}
\end{equation}
where $t$ denotes the current training environment steps and $T_{tighten}$ represents the tightening period. In the initial stages of training, $S$ remains close to 1.0, providing a relatively tolerant exploration space. As training progresses, $S$ gradually converges to $S_{final}$ (set to 0.7), causing the position threshold (proportional to $S^3$) and the fingertip threshold (proportional to $S$) to tighten progressively, thereby forcing the policy to achieve higher manipulation precision in the later stages.

Finally, during the inference and evaluation phase, to standardize the assessment of manipulation capabilities and ensure fair benchmarking, we adopt fixed physical criteria for success determination. A test episode is classified as successful if and only if the object's state deviation remains strictly within safe thresholds at every time step $t$: specifically, the object position error must satisfy $\|p_{obj} - p_{obj}^E\|_2 \le 3 \text{ cm}$, and the rotational error must satisfy $\theta_{diff} < 30^{\circ}$. Any violation of these conditions at any point during the operation results in the immediate classification of the episode as a failure.

\section{Experimental Details}
\label{sec:ed}
\subsection{Experimental Parameter Setting}
\label{sec:pa}
To supplement the configuration, this section details the specific hyperparameters for the PPO optimizer, the initial gains for the \textit{shadow engine}, and the specifications of the robotic hands used in our experiments.
\subsubsection{Optimization and Network Hyperparameters}
We utilize the PPO algorithm for policy learning. The specific settings for gradient clipping, entropy regularization, and batch updates are listed in \cref{tab:ppo_params}.
\begin{table}[h]
\centering
\caption{PPO Hyperparameters and Training Configuration.}
\label{tab:ppo_params}
\small
\begin{tabular}{lc} 
\toprule
\textbf{Parameter} & \textbf{Value} \\ 
\midrule
Learning Rate & $5 \times 10^{-4}$ \\
Mini-batch Size & $1,024$ \\
Horizon Length & $32$ \\
Optimization Epochs  & $5$ \\
Discount Factor & $0.99$ \\
GAE Parameter & $0.95$ \\
Clip Range& $0.2$ \\
Entropy Coefficient ($c_{\text{start}}^{\text{entropy}}$) & $5 \times 10^{-4}$ \\
Value Loss Coefficient ($c_{vf}$) & $4.0$ \\
Bound Loss Coefficient ($c_{bound}$) & $0.1$ \\
Max Gradient Norm & $1.0$ \\
\bottomrule
\end{tabular}
\end{table}
\subsubsection{Dynamic Control and Curriculum Parameters}
\cref{tab:dynamic_params} presents the specific values for the dynamic components of UniBYD, including the reference object mass and baseline PD gains for the \textit{shadow engine}, the weight extrema for both imitation and goal rewards, and the sliding window size utilized for metric computation.
\begin{table}[h]
\centering
\caption{Additional Dynamic Control and Curriculum Parameters.}
\label{tab:dynamic_params}
\small
\begin{tabular}{lc} 
\toprule
\textbf{Parameter} & \textbf{Value} \\
\midrule
Reference Object Mass ($m_{ref}$) & 0.03788 \\
Baseline Prop. Gain ($K_{p,cfg}$) & 10.0 \\
Baseline Deriv. Gain ($K_{d,cfg}$) & 3.0 \\ 
Max Imitation Weight ($w_{max}^{imi}$) & 1.0 \\
Min Imitation Weight ($w_{min}^{imi}$) & 0.2 \\
Max Goal Weight ($w_{max}^{goal}$) & 3.0 \\
Min Goal Weight ($w_{min}^{goal}$) & 0.2 \\ 
Sliding Window Size ($M$) & 100 \\
\bottomrule
\end{tabular}
\end{table}
\subsubsection{Unified Morphological Representation and Hand Specifications}
To enable consistent modeling across diverse embodiments, we set the maximum number of joint degrees of freedom for UMR to $D_{max} = 22$. This dimension is sufficient to accommodate the vast majority of existing robotic hands. For hands with fewer degrees of freedom, zero-padding is applied. The specific robotic hands supported and their corresponding degrees of freedom are listed in \cref{tab:hand_specs}.
\begin{table}[h]
\centering
\caption{Specifications of Supported Robotic Hands.}
\label{tab:hand_specs}
\small
\begin{tabular}{lc} 
\toprule
\textbf{Robotic Hand} & \textbf{Degrees of Freedom (DOF)} \\ 
\midrule
Shadow Hand & 22 \\
Allegro Hand & 16 \\
Inspire Hand & 6 \\
OHand$^{TM}$ & 11 \\
CasiaHand 3-Finger & 10 \\
XArm Gripper & 1 \\
Franka Panda & 1 \\
\bottomrule
\end{tabular}
\end{table}

\subsection{Motion Retargeting}
To bridge the significant embodiment gap between the human hand (represented by the MANO model) and heterogeneous robotic hands, we employ an optimization-based inverse kinematics method. This approach seeks to identify the optimal robotic joint configuration $\mathbf{q}$ and the 6-DoF pose of the robotic wrist $\mathcal{P}_{wrist}$ such that the robot's end-effectors spatially align with the human expert's demonstration in Cartesian space.

We formulate the retargeting process for each step $t$ as a non-linear least squares optimization problem. The decision variables include the wrist position $\mathbf{p}_{wrist} \in \mathbb{R}^3$, the wrist rotation parameters $\mathbf{r}_{wrist} \in \mathbb{R}^6$ (using a continuous 6D rotation representation), and the robot's joint angle vector $\mathbf{q} \in \mathbb{R}^{D_h}$. The objective function $J$ is defined as the weighted Euclidean distance between the corresponding keypoints:
\begin{equation}
\begin{split}
  & J(\mathbf{p}_{wrist}, \mathbf{r}_{wrist}, \mathbf{q}) = \frac{1}{N_{kp}} \sum_{i=1}^{N_{kp}} w_i \cdot \\
  & \left\| \Phi_i(\mathbf{p}_{wrist}, \mathbf{r}_{wrist}, \mathbf{q}) - \mathbf{x}_i^{human} \right\|_2 ,
\end{split},
\label{eq:important}
\end{equation}
where $N_{kp}$ denotes the total number of matched keypoints. $\mathbf{x}_i^{human} \in \mathbb{R}^3$ represents the 3D position of the $i$-th human keypoint (fingertip or joint center) derived from the MANO model. $\Phi_i(\cdot)$ is the differentiable forward kinematics function that computes the Cartesian coordinates of the robot's $i$-th keypoint under the current configuration. $w_i$ is a scalar weight assigned to the $i$-th keypoint to regulate its influence on the optimization.

Given the diverse kinematic structures of robotic hands, we automatically establish correspondences between robot links and human joints based on semantic naming conventions in the URDF. To prioritize the precision of manipulation contacts, we implement a hierarchical weighting strategy. As the primary contact interfaces, fingertips are assigned the highest weights $w_{tip} \in [20, 30]$. Specifically, for 5-finger dexterous hands, the thumb and index fingertips are often weighted between 25 and 30 to ensure the faithful reproduction of fine pinching motions. To maintain a natural hand pose and prevent non-physical contortions, intermediate links and the wrist base serve as auxiliary constraints with significantly lower weights, typically $w_{link} \approx 1$ to $5$. This non-uniform weighting ensures that the optimizer converges primarily on the fingertip positions while utilizing the null space to maintain a plausible overall posture.

To ensure the physical feasibility of the retargeted motion, we enforce the following constraints and mechanisms during optimization:
\begin{itemize}
    \item \textbf{Joint Limits:} We impose strict constraints on the joint angles using a hard clamping function to keep predictions within the physical range defined by the URDF:
    \begin{equation}
    \mathbf{q}_{clamped} = \text{clamp}(\mathbf{q}, \mathbf{q}_{lower}, \mathbf{q}_{upper}),
    \label{eq:joint_clamp}
    \end{equation}
    where $\mathbf{q}_{lower}$ and $\mathbf{q}_{upper}$ are the lower and upper joint limits, respectively.
    \item \textbf{Mimic Joint Constraints:} For underactuated grippers with mechanically coupled joints, we explicitly enforce coupling logic prior to the forward kinematics computation. For instance, slave joint values are forced to match the master joint ($\mathbf{q}_{slave} = \mathbf{q}_{master}$) to reflect the actual transmission mechanics.
    \item \textbf{Solver Configuration:} We utilize the Adam optimizer for iterative solving. The learning rates are set to $8 \times 10^{-4}$ for the wrist pose ($\mathbf{p}_{wrist}, \mathbf{r}_{wrist}$) and $4 \times 10^{-4}$ for the joint angles $\mathbf{q}$. The optimization is parallelized on the GPU, allowing for a maximum of 4,000 iterations per frame, with an early stopping mechanism triggered if the loss improvement falls below a threshold $\epsilon = 10^{-5}$.
\end{itemize}

Crucially, the optimization-based retargeting method described above serves a dual purpose in this study. Firstly, it constitutes the specific implementation detail for the Retargeting baseline compared in the experimental section. Secondly, for UniBYD, this method is utilized to generate the physically feasible initial poses for the robotic hand. These poses are used to initialize the environment at the beginning of each training episode, ensuring that the reinforcement learning policy starts from a valid state close to the object before beginning its autonomous exploration.

\subsection{ManipTrans and Reproduction of DexMachina}
\subsubsection{ManipTrans}
To rigorously evaluate the effectiveness of UniBYD, we conduct comparative experiments against ManipTrans, which represents the current state-of-the-art method for dexterous manipulation. To ensure the fairness and uniformity of the comparison, ManipTrans is implemented using the exact same experimental configuration as UniBYD, including identical hardware equipment, the same component weights for the imitation reward, and the same strict testing standards (PE $\le 3$ cm and OE $< 30^{\circ}$).

For the 5-fingered setting, we follow the original ManipTrans setup by using the OakInk-V2 dataset, which is also used to pre-train its released open-source imitator. We do not retrain its first stage from scratch for each specific task. Instead, we directly use the official general checkpoint released by ManipTrans for the first stage. This protocol preserves the intended generalization ability of ManipTrans and provides the fairest available baseline comparison.

\subsubsection{Reproduction of DexMachina}
As a task-centric approach, DexMachina's core advantage lies in introducing the Virtual Object Controller Curriculum and a unique reward structure. We reproduce DexMachina's key mechanisms within the unified environment of UniBYD.

We first reproduce the Virtual Object Controller Curriculum. DexMachina's core idea is to use virtual object controllers with decaying strength to drive the object, allowing the policy to learn under guided conditions before eventually taking over control. To replicate this mechanism, we implement PD controllers with gain annealing. The controller is initialized with proportional gain and derivative gain, and by setting a decay schedule, it accurately replicates the virtual object controller with decaying strength described in the original paper.

The second step is the faithful reproduction of DexMachina's reward functions. DexMachina's Task Reward is its primary learning signal, which adopts a multiplicative structure computed as the product of exponential terms over the object's position error, rotation error, and joint-angle error. In addition, DexMachina uses auxiliary rewards, including the Motion Imitation Reward and Contact Reward. In our implementation, we strictly follow DexMachina's official reward structure and precisely reproduce these reward terms. The reproduced baseline is faithful to DexMachina's task-centric and exploration-prioritized design.

To ensure the fairness and validity of the experimental comparison, we use the exact same hardware devices, the same early stopping strategy, and the same testing criteria as UniBYD during the reproduction of DexMachina.

\subsection{Early Stopping Strategy}
We employ an early stopping strategy based on multi-epoch-scale slope detection to robustly determine when training has reached a convergence plateau, differing from traditional methods that rely on monitoring validation loss patience. This strategy aims to prevent training from stopping prematurely or consuming computational resources inefficiently by analyzing the growth trend of the average reward curve across different epoch windows.

The core decision for early stopping is based on the history of the average reward recorded after each training epoch. We use the linear regression to quantify the slope $\text{Slope}_W$ of the reward as it changes with epochs. For a given epoch window $W$, the slope is calculated as:
\begin{equation}
\text{Slope}_W = \text{linregress}(e, \mathbf{R}_W)[0],
\label{eq:joint_clamp}
\end{equation}
where $e$ is the epoch index sequence, and $\mathbf{R}_W$ is the sequence of average reward values from the most recent $W$ epochs. $\text{Slope}_W$ represents the average growth rate of the reward over $W$ epochs.

To ensure the robustness of the convergence judgment, we simultaneously monitor the slope trends across three different epoch windows $W$: short-term ($W=8$ epochs), medium-term ($W=32$ epochs), and long-term ($W=64$ epochs). This mechanism is activated only after the number of training epochs exceeds a minimum detection threshold,  $\delta_{\text{early}} = 150$.

Early stopping is triggered only if the reward slope across all three epoch windows $W$ is less than a small, predefined threshold $\delta_{\text{slope}} = 1 \times 10^{-4}$. Training terminates if and only if the following compound logical condition is met:
\begin{equation}
\label{eq:early_stop_criteria}
\begin{split}
\text{EarlyStop} = \mathbb{I} \bigg[ (\text{Epoch} \ge \delta_{\text{early}} \quad \land\\
 \left. [\bigwedge_{W \in \{8, 32, 64\}} (\text{Slope}_W \le 1 \times 10^{-4})] \right].
\end{split}
\end{equation}

This multi-scale coupled condition effectively prevents false positives caused by local fluctuations or noise in the reward curve, ensuring that the policy has fully exploited its performance potential before training is halted.

\subsection{Real-World Experiment Details}
\subsubsection{Detailed Deployment Pipeline}
This is a hierarchically decoupled engineering system. First, we use a RealSense D435i camera and the FoundationPose algorithm to perform object-centric 6D pose estimation. Second, we need to drive the robotic arm to an initial pose identical to that in the simulator. We do not use UniBYD to control the robotic arm; instead, we adopt classical planning. Based on the detected object pose and the relative hand-object pose in the simulator, combined with pre-calibrated hand-flange extrinsics, we calculate the global target for the arm's end-effector. Subsequently, we utilize IK and the RRT-Connect planner (MoveIt!) to transport the dexterous hand collision-free to the initial pose. Finally, once the arm is in position, UniBYD takes over the manipulation prediction of the dexterous hand. At this stage, UniBYD operates at a frequency of 20Hz, and the system utilizes the aforementioned extrinsics to calculate the precise pose of the object relative to the wrist in real-time, constructing the relative observation space.

\subsubsection{Failure Analysis \& Zero-Shot Transfer Mechanism}

\textbf{a. Mechanism of Zero-Shot Transfer} 

(1) \textbf{State Consistency via Initialization:} We utilize robotic arm planning to physically transport the dexterous hand to a Relative Initial Pose exactly consistent with the simulator.

(2) \textbf{Constraint via Boundary Loss:} The Boundary Loss introduced during training forces the policy to output actions that remain within safe physical joint limits. This embedded constraint makes the policy inherently safe on the real robot, preventing extreme command outputs.

\textbf{b. Sim-to-Real Gap}

(1) \textbf{Independence from SE:} The \textit{shadow engine} (SE) serves solely as a training aid (and is fully annealed before training ends); it does not exist during inference or real-world deployment. Thus, failures in real-world experiments are unrelated to SE removal.

(2) \textbf{Sim-to-Real Performance Gap:} Our experiments indicate a performance loss. The tasks in the paper achieve success rates exceeding 95\% in simulation, whereas the average success rate on the real robot is 62\%.

\textbf{c. Detailed Failure Analysis}

We compile detailed statistics on the specific causes of task failure across 150 trials:

(1) \textbf{Hand-Object Collision (~19\%):} This is the primary source of failure. Unintended collisions occur during dexterous manipulation, causing the object to be knocked over or displaced before a grasp is established.

(2) \textbf{Object Drops (~9\%):} During lifting or reorientation, the object slips due to an insecure grip.

(3) \textbf{Joint/Workspace Limit (~7\%):} Robot self-collision almost never occurs. Such failures mainly happen because the arm or dexterous hand reaches joint limits when the policy attempts to track certain trajectories, triggering low-level emergency stops.

(4) \textbf{Robot-Table Collision (~3\%):} Minor collisions occur between the robot fingertips or the object and the workbench during manipulation, causing the robot to emergency stop.

\section{Further Implementation Details of the UniManip Benchmark}
\label{sec:uni}
\subsection{Generating Expert Data for Diverse Robotic Morphologies from Human Hand Demonstrations}
Due to the immense morphological discrepancy between human and diverse robotic hands, human demonstrations cannot directly serve as expert data for robotic hands of diverse configurations, particularly 2- and 3-fingered manipulators. To overcome this problem, we designed an iterative retargeting pipeline based on a Multimodal Large Language Model (MLLM) for generating high-quality cross-morphology expert data. We utilize the Gemini 2.5 Pro model as the core MLLM engine.
\subsubsection{Rigid Body Mapping Generation}
The model receives the raw Mocap human hand motion data and the target robot's URDF file. The model's core task is to directly establish a one-to-one functional mapping ($\mathcal{M}_{rigid}$) between the human hand rigid bodies and the robotic hand rigid bodies.

This mapping must possess functional equivalence. Every critical rigid body of the robotic hand, including all fingertips and major metacarpophalangeal joints, must find a corresponding human rigid body on the human hand model that performs an equivalent kinematic function. The MLLM must rely on its understanding of the robot's topological structure and kinematic function to establish this mapping that bridges the morphological gap. The specific prompt used is shown below:

\begin{tcolorbox}[breakable, colback=gray!5, colframe=gray!50, boxrule=0.5pt, left=2mm, right=2mm]
[ROLE AND MISSION DEFINITION]

You are a senior robotics kinematics expert and a multimodal structure analysis engine. Your primary mission is to generate a high-precision, high-functional equivalence rigid body mapping table ($\mathcal{M}_{rigid}$) based on the provided human motion data and the target robot's URDF file.

[INPUT DATA STRUCTURE]
\begin{enumerate}
    \item Human Hand Structure (Source - MANO):
        \begin{itemize}
            \item List of human keypoint names (e.g., thumb\_tip, index\_mcp, wrist).
            \item Description of human hand topological structure (joint parent-child relationships, to aid functional understanding).
        \end{itemize}
    \item Robot Hand Structure (Target - URDF Parse):
        \begin{itemize}
            \item List of all rigid body names for the target robot (e.g., rh\_ff\_tip\_link, panda\_leftfinger).
            \item Description of human hand topological structure (joint parent-child relationships, to aid functional understanding).
        \end{itemize}
\end{enumerate}

[CORE TASK AND MAPPING REQUIREMENTS]

Generate a high-precision, one-to-one functional mapping table from the robot rigid bodies to human keypoints.
\begin{enumerate}
    \item Functional Equivalence ($\mathcal{M}_{rigid}$): The mapping must ensure functional equivalence. Specifically, every link of the robot must be mapped to a corresponding human rigid body that possesses the same kinematic function.
    \item Completeness and Criticality: The mapping must include all critical functional links deemed essential for grasping, particularly the correspondence for all Fingertips and major Proximal Joints.
    \item Kinematic Plausibility: When establishing the mapping, you must rely on an understanding of the robot's topological structure to ensure the mapping does not introduce kinematic conflicts (e.g., the robot's distal link must not be mapped to the human hand's proximal joint).
\end{enumerate}

[OUTPUT FORMAT SPECIFICATION]

Strictly return the mapping table ($\mathcal{M}_{rigid}$) in JSON format. The Key must be the robot rigid body name, and the value must be the human keypoint name.

\{

    \quad ``robot\_rigid\_1\_name": ``human\_keypoint\_name\_A",
    
    \quad ``robot\_rigid\_2\_name": ``human\_keypoint\_name\_B",
    
    \quad ...
    
\}

\end{tcolorbox}

The MLLM subsequently performs a rigorous internal validation of the generated mapping $\mathcal{M}_{rigid}$, ensuring its compliance with advanced kinematic and geometric constraints. This validation is a critical safeguard before proceeding to simulator retargeting. This process involves three core checks:

Firstly, the kinematic plausibility check requires the MLLM to verify the topological structure of the mapping, preventing non-physical correspondences. For example, a robot's distal link must not be mapped to the human hand's proximal joint, thereby preserving the topological order and relative motion relationships of the kinematic chain. Secondly, during the geometric and DOF compatibility check, the MLLM utilizes its URDF parsing capability to assess whether the mapping is compatible with the hand's intrinsic degrees of freedom and joint types (such as revolute versus prismatic joints), thus ensuring the generated trajectory is physically feasible. Finally, the initial feasibility heuristics check involves the MLLM performing a preliminary assessment, based on its understanding of the initial relative positions of the object and hand, to determine if the mapping would immediately cause obvious interpenetration or unreachability in the very first frame. If the validation fails, the MLLM automatically regenerates the mapping until the internal consistency check is passed. The specific prompt is shown below:

\begin{tcolorbox}[breakable, colback=gray!5, colframe=gray!50, boxrule=0.5pt, left=2mm, right=2mm]

[ROLE AND MISSION DEFINITION]

You are a Senior Robotics Kinematics Validation Expert and a Topological Structure Analysis Engine. Your primary mission is to perform a rigorous \textbf{Three-Stage Internal Consistency Check} on the newly generated rigid body mapping table ($\mathcal{M}_{rigid}$) to determine its physical plausibility and kinematic compliance before it proceeds to simulator retargeting.

[INPUT DATA]

\begin{enumerate}
    \item \textbf{Mapping Table Under Validation:} $\mathcal{M}_{rigid}$ (JSON format).
    \item \textbf{Robot Structure Data:} The URDF topological structure (parent-child relationships, link list) and the DOF list (including joint types: Revolute/Prismatic).
    \item \textbf{Initial State Data:} The initial relative position and orientation of the human hand and the target object at step $t=0$.
\end{enumerate}

[THREE-STAGE CHECK REQUIREMENTS]

The mapping $\mathcal{M}_{rigid}$ must pass all three comprehensive checks:

\begin{enumerate}
    \item \textbf{Kinematic Plausibility Check}
    \begin{itemize}
        \item \textbf{Requirement:} Verify the topological sequence of the mapping to prevent non-physical correspondences. For instance, a robot's distal link \textbf{must not} be mapped to the human hand's proximal joint.
        \item \textbf{Goal:} Ensure the topological order and relative motion relationships of the kinematic chain are maintained.
    \end{itemize}

    \item \textbf{Geometric and DOF Compatibility Check}
    \begin{itemize}
        \item \textbf{Requirement:} Utilize the URDF file analysis to assess whether the mapping is compatible with the robot's intrinsic Degrees of Freedom (DOF) count and specific joint types (e.g., a revolute joint cannot map to an inappropriate linear motion).
        \item \textbf{Goal:} Ensure the resulting trajectory is physically executable.
    \end{itemize}

    \item \textbf{Initial Feasibility Heuristic Check}
    \begin{itemize}
        \item \textbf{Requirement:} Based on the initial state data, perform a preliminary assessment of the mapping. Determine if the mapping would lead to obvious \textbf{interpenetration} (self-collision or object penetration) or \textbf{unreachability} in the very first frame.
        \item \textbf{Goal:} Filter out guaranteed failure mappings before computationally expensive simulation.
    \end{itemize}
\end{enumerate}

[OUTPUT FORMAT SPECIFICATION]

Strictly return a JSON object containing the \textbf{final verdict} and a \textbf{detailed reasoning chain}.

\{

    \hspace*{2em}``verdict": ``GO" \textbar  ``REGENERATE", \\
    \hspace*{2em}``reasoning\_chain": ``Detailed logic explaining the successful passage of all three checks, or specifying the exact reason for failure (e.g., Kinematic check failed because robot\_link\_X maps proximal joint to distal joint).", \\
    \hspace*{2em}``failed\_check\_type": ``None" \textbar ``Kinematic" \textbar ``Geometric" \textbar ``Feasibility"

 \}
\end{tcolorbox}

\subsubsection{Simulation Retargeting and Iterative Evaluation}
Once the mapping $\mathcal{M}_{rigid}$ passes the MLLM's internal validation, it is applied in the simulator to drive the robotic hand's retargeting execution. Subsequently, we initiate an iterative feedback loop, utilizing the Gemini 2.5 Pro model as a high-level visual expert evaluator to guarantee the quality and morphological adaptability of the retargeted trajectory.
\begin{enumerate}
    \item Visual Data Acquisition and Multimodal Input: After each retargeting execution, to provide comprehensive visual evidence, we capture images of the current robotic hand pose from four standardized viewpoints (front, back, left, and right). This visual input is packaged along with the contextual information from the original Mocap data and provided as input for the MLLM's evaluation.
    \item MLLM Morphological Adaptability Judgment: The Gemini 2.5 Pro model combines its prior understanding of the robot's structure (derived from the URDF) with the visual information in the images to perform a high-level assessment of the retargeted pose. The core of the judgment focuses on evaluating the embodied appropriateness and visual quality of the pose. The model must infer whether the current pose fully leverages the target robotic hand's morphological characteristics and whether there is any obvious interpenetration or unstable grasping configuration.
    \item Feedback and Termination Mechanism: If the MLLM determines that the current pose is visually and functionally unacceptable, the model then automatically regenerates a new set of rigid body mappings ($\mathcal{M}_{rigid}'$), driving the simulator into the next round of IK optimization iteration. This iterative feedback mechanism, centered on the MLLM, ensures that the final output expert data is not only kinematically feasible but also highly adapted to the target robotic hand's morphology in both visual and functional terms, supporting the subsequent UniBYD training.
\end{enumerate}

The specific prompt used is shown below:

\begin{tcolorbox}[breakable, colback=gray!5, colframe=gray!50, boxrule=0.5pt, left=2mm, right=2mm]

[ROLE AND MISSION DEFINITION]

You are a \textbf{High-Level Visual Expert Evaluator} and a kinematics reasoning engine. Your primary task is to perform a functional and visual validation of the current retargeted robotic hand pose against the human intent and the robot's mechanical structure.

[INPUT DATA STREAMS]

\begin{enumerate}
    \item \textbf{Visual Evidence (4 Images):}  
Four standardized viewpoint images of the current robot pose interacting with the object (Front, Back, Left, Right).

    \item \textbf{Structural Context:}  
Prior knowledge of the \textbf{Robot's Morphological Characteristics} (derived from URDF analysis).
    \item \textbf{Functional Context:}  
The original \textbf{Human Mocap Intent} (e.g., target object pose, action type, desired contact points).
\end{enumerate}

[THREE-POINT JUDGMENT CRITERIA]
Perform a comprehensive, high-level assessment of the pose quality. \textbf{A verdict of 'REGENERATE' must be returned if any criterion fails.}

\begin{enumerate}
    \item \textbf{Embodied Appropriateness (Functional Feasibility)}
    \begin{itemize}
        \item \textbf{Check:}  
Does the pose successfully achieve the functional objective dictated by the human demonstration (e.g., stable pinching, enclosing) while \textbf{strictly adhering to the robot's physical and kinematic capability}? The pose must reflect a viable path to the intended grasp configuration.

        \item \textbf{Failure Example:}  
A multi-finger hand, which possesses the mechanical capability to form a stable full-web-space wrap, defaults to an inefficient or unstable minimalist action (e.g., only two fingers for a large object). The pose must utilize the robot's capacity to maximize functional stability.
    \end{itemize}

    \item \textbf{Visual Quality and Physical Plausibility}
    \begin{itemize}
        \item \textbf{Check:}  
Is the pose physically sound and visually stable?

        \item \textbf{Failure Examples:}
(a) Obvious Interpenetration (self-collision or object-link penetration). (b) Fingertips are clearly forming an unstable grasping configuration (e.g., near or past the object center point, visually indicating slippage).

    \end{itemize}

    \item \textbf{Grasp Stability Heuristics}
    \begin{itemize}
        \item \textbf{Check:}  
              Based on the visual evidence, are the fingers forming clear opposing forces or are they merely resting against the object?

        \item \textbf{Failure Example:}  
The fingers are resting against the object instead of forming a clear \textbf{opposing force} required for a stable grasp in this context.
    \end{itemize}
\end{enumerate}

[OUTPUT FORMAT SPECIFICATION]

Strictly return a JSON object with the final verdict and reasoning. The output is used by the external control script to determine if a new mapping ($\mathcal{M}_{rigid}'$) must be generated.

\{

\hspace*{2em}``verdict'': ``ACCEPT'' \textbar{} ``REGENERATE'', \\
\hspace*{2em}``failed\_criteria": [``None'', ``Embodied Appropriateness'', ``Visual Quality'', ``Grasp Stability''], \\
\hspace*{2em}``reasoning\_summary": ``Provide a brief explanation (e.g., `Failure: Obvious self-collision observed between the ring finger and the palm.', or `Success: Pose is efficient and stable, utilizing the full three-finger web space.')''

\}

\end{tcolorbox}

\subsection{Data Category Distribution}
As shown on the left of \cref{fig:d}, the data categories of the UniManip benchmark are primarily divided into four major task types: two-finger unimanual, three-finger unimanual, five-finger unimanual, and five-finger bimanual. These are further subdivided into 29 total task categories.

\subsection{Details of AS Computation}

The MLLM and 100 human volunteers both score the robotic manipulation strategy based on the identical evaluation prompt. We utilize the MLLM to score the same operation ten times using the identical prompt. The final AS metric is then derived by averaging these ten model scores together with the scores provided by the 100 human volunteers. The specific prompt used is shown below:

\begin{figure*}[t]
  \centering
  \includegraphics[width=\linewidth]{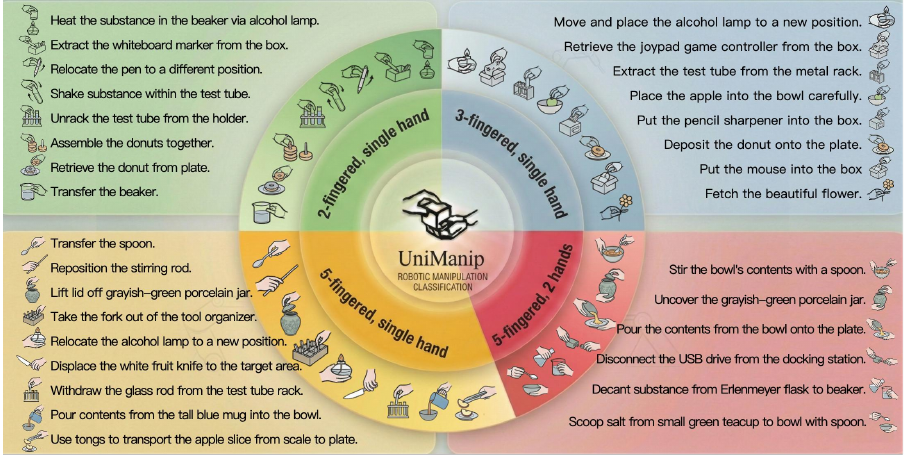}
   \caption{The distribution of UniManip.}
   \label{fig:d}
\end{figure*}

\begin{figure*}[t]
  \centering
  \includegraphics[width=\textwidth]{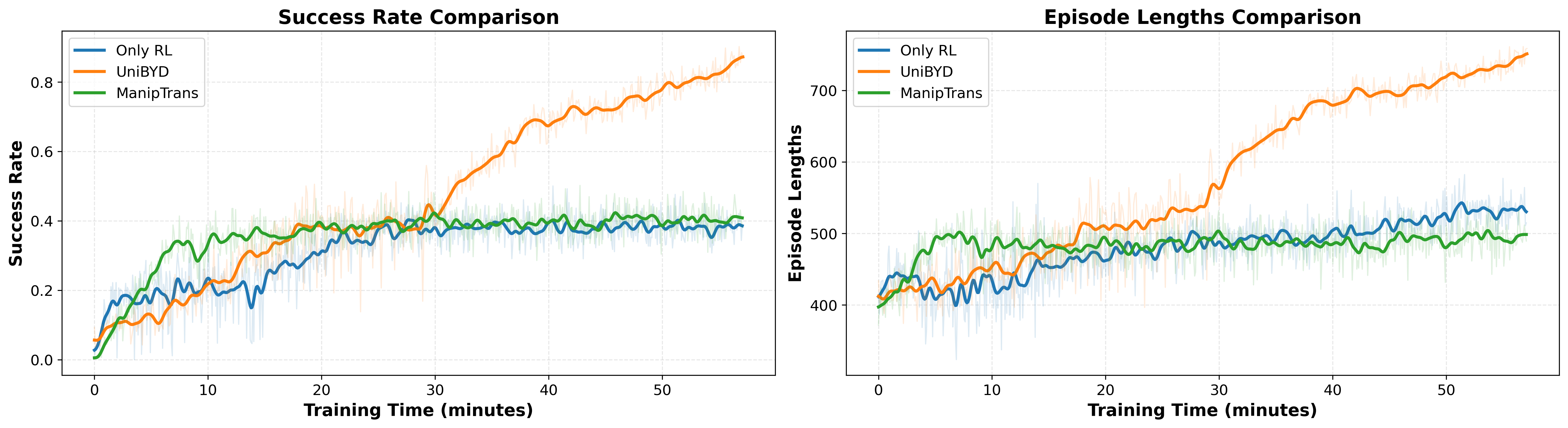}
   \caption{The evolution of success rate and episode length over training time for a representative task: Pouring liquid from a tall blue mug into a bowl.}
   
   \label{fig:ex2}
\end{figure*}

\begin{tcolorbox}[breakable, colback=gray!5, colframe=gray!50, boxrule=0.5pt, left=2mm, right=2mm]

[TASK]

You are an expert in dexterous robotic manipulation and embodiment adaptation. You are given four sequential images showing the manipulation process of a robotic hand performing a specific task. The robotic hand may have different morphologies (e.g., two-finger, three-finger, or five-finger).

Your goal is to carefully analyze the manipulation strategy shown in the images and assess how well the operation method fits the hardware characteristics of the robotic hand.

[EVALUATION FOCUS]

\begin{enumerate}
    \item \textbf{Embodiment Adaptation (Hardware Compatibility):}
    \begin{itemize}
        \item Does the manipulation strategy fully leverage the mechanical structure, degrees of freedom, and motion range of this robotic hand?
        \item Are the contact patterns, grasp poses, and motion trajectories appropriate for this specific morphology?
        \item Is the control strategy (e.g., parallel pinch, rolling, regrasping, finger coordination) mechanically feasible and efficient for this hardware?
    \end{itemize}

    \item \textbf{Manipulation Quality (Effectiveness \& Naturalness):}
    \begin{itemize}
        \item Is the manipulation stable, efficient, and smooth across the sequence?
        \item Does the sequence demonstrate coordinated control and realistic grasp transitions?
        \item Are there unnecessary or suboptimal motions that reduce effectiveness?
    \end{itemize}
\end{enumerate}

[SCORING]

Give a \textbf{single numerical score from 0 to 10}, where:

\begin{itemize}
    \item \textbf{0--2:} The manipulation strategy is unsuitable for this hardware and performs poorly.
    \item \textbf{3--5:} The manipulation is partially feasible but inefficient or poorly adapted.
    \item \textbf{6--8:} The manipulation fits the hardware well and is mostly effective.
    \item \textbf{9--10:} The manipulation is excellently adapted to the hardware, showing highly efficient, natural, and optimal control.
\end{itemize}

[OUTPUT FORMAT]
\begin{verbatim}
Score: X/10
Explanation: [Brief, objective
reasoning focusing on 
adaptation and quality]
\end{verbatim}
\end{tcolorbox}

\section{Performance Evolution and Episode Length Variation over Training Time}
\label{sec:pe}
We compare the proposed UniBYD framework against the imitation-driven baseline ManipTrans and the pure exploration baseline Only RL, to analyze the evolution of SR and executed Episode Length over time during training. The Only RL strategy uses UniBYD's reward settings but deliberately removes the imitation reward component for robotic hand movement.

As shown on the left of \cref{fig:ex2}, the training results clearly demonstrate the impact and limitations of the imitation reward on learning efficiency. Due to the strong guidance of the dense imitation reward, ManipTrans's success rate rapidly increases during the early training phase (0-25 minutes). However, this strict imitation quickly locks the policy into a local optimum, causing its success rate to rapidly converge and stagnate around 0.4, failing to achieve further breakthroughs. In contrast, the Only RL strategy shows a slower initial increase in SR because it lacks the guidance of the imitation reward, stabilizing similarly around 0.4. This result indicates that relying purely on sparse goal rewards for exploration makes it difficult to effectively learn the initial configurations required for complex operations.

The UniBYD curve (orange) highlights the advantages of the dynamic reinforcement learning mechanism. While UniBYD's SR is not the highest initially because it deliberately maintains high entropy for exploration, it rapidly converges to a success rate level similar to ManipTrans (around 0.4). The critical breakthrough occurs in the mid-to-late phase of training (after approximately 30 minutes). Supported by the Goal Reward ($R^{goal}$) and sustained exploration entropy, UniBYD successfully breaks through the local optimum set by the Imitation Reward. The policy explores manipulation methods better suited to the robot's physical morphology, and the SR subsequently leaps, ultimately stabilizing around 0.8, significantly outperforming other baselines. The corresponding Episode Length trend mirrors this success rate, demonstrating that UniBYD not only achieves higher SR but also learns more stable, longer, and thus more reliable trajectories.

As shown on the right side of \cref{fig:ex2}, the episode length curves for ManipTrans and Only RL quickly reach a plateau in the early training phase, stabilizing around 500 steps. This confirms that although these two baselines can maintain a success rate of around 0.4, this success is mainly achieved in shorter, end-of-trajectory segments. This phenomenon is due to our training process randomly selecting a step within the full Episode to start training. This also explains why the success rate of these two baselines in real testing (where the task starts from the beginning) is far lower than during training, as their policies fail to maintain stability in longer task sequences.

In contrast, UniBYD's episode length grows synchronously and significantly with the rapid rise in its success rate, ultimately breaking through 750 steps. This strongly demonstrates UniBYD's ability to complete the full, long-horizon task with high robustness, rather than relying on imitation of segments. Furthermore, we observe an interesting phenomenon: although the success rate of Only RL remains slightly lower than ManipTrans until the end of training, its episode length continues to increase and gradually surpasses ManipTrans. This potentially means that if training is to continue, the pure exploration nature of the Only RL policy possesses the potential to surpass the final performance of the imitation-driven ManipTrans.

\section{Algorithm of UniBYD}
\label{sec:al}
\cref{alg:unibyd_framework} summarizes the procedure of the UniBYD framework.

\begin{algorithm}
\caption{UniBYD Framework}
\label{alg:unibyd_framework}
\begin{algorithmic}[1]
\REQUIRE Expert Demonstrations $\mathcal{D}_E$, Robotic Hand Set $\mathcal{H}$, Max Epochs $E_{max}$.
\STATE Initialize Policy $\pi_{\theta}$, Value Function $V_{\phi}$, Sliding Window Size $M$.
\STATE Initialize Thresholds: $T_{decay}$, $T_{early}$, $\delta_{SR}$, $\delta_{min}$ (0.2), and Scaling Factor $\gamma_{scale}$.
\STATE Initialize Baseline Gains $(K_{p,cfg}, K_{d,cfg})$ and Reference Mass $m_{ref}$.
\FOR{epoch $e = 1$ to $E_{max}$}
    \STATE // \textit{1. Parameter and Reward Weight Updates}
    \STATE Compute Hand Blending Weight $\beta_t \leftarrow \max(0, 1 - e/T_{decay})$.
    \STATE Calculate Initial Gains: $K_p^{start}, K_d^{start} \leftarrow f(m_{obj}, m_{ref}, K_{p,cfg}, K_{d,cfg})$.
    \STATE Update Object PD Gains: $K_{p,e}, K_{d,e} \leftarrow K^{start} \cdot \beta_t$.
    \STATE Compute Recent Success Rate $\bar{SR}$ using window $M$.
    \STATE Compute Transition Factor $f \leftarrow \max(0, \min(1, (\bar{SR} - \delta_{min}) \times \gamma_{scale}))$.
    \STATE Update $w_{e}^{imi}$ and $w_{e}^{goal}$ using piecewise formulations based on $e$ and $\bar{SR}$.

    \STATE // \textit{2. Data Collection (Hybrid MDP with Shadow Engine)}
    \FOR{step $t = 1$ to $T_{horizon}$}
        \STATE Get observation $o_t$ using UMR.
        \STATE Compute Model Action $\Delta a_{t}^{\pi} \leftarrow \pi_{\theta}(o_{t})$ and retrieve Expert Action $\Delta a_{t}^{E}$.
        \STATE // \textit{Execute Shadow Engine Guidance}
        \STATE $\Delta a_{t}^{exec} \leftarrow (1-\beta_t) \Delta a_{t}^{\pi} + \beta_t \Delta a_{t}^{E}$.
        \STATE Apply Object Support Force $F_{support}$ via PD controller $(K_{p,e}, K_{d,e})$.
        \STATE Execute $\Delta a_{t}^{exec}$, observe $s_{t+1}$, and determine episodic success $\mathcal{C}_{suc}$.
        \STATE Store transition $(o_t, \Delta a_{t}^{exec}, R_{t}^{imitation}, \dots)$.
    \ENDFOR

    \STATE // \textit{3. Policy Update (Dynamic PPO Loss Synergy)}
    \STATE Compute Total Reward $R_{t} \leftarrow w_{e}^{imi} R_{t}^{imitation} + w_{e}^{goal} R^{goal}$.
    \STATE Compute PPO loss $L_{t}(\theta)$ including Entropy and Bound Loss.
    \STATE Update policy parameters: $\theta \leftarrow \theta - \eta \nabla_{\theta} L_{t}(\theta)$.
\ENDFOR
\end{algorithmic}
\end{algorithm}

\section{Distinct Manipulation Policies of Different Robotic Hands for the Same Task}
\label{sec:dis}
For the same task (Assemble the donuts together), UniBYD is able to discover manipulation strategies tailored to the physical characteristics of different robotic hands.

\begin{figure}[t]
  \centering
  \includegraphics[width=\linewidth]{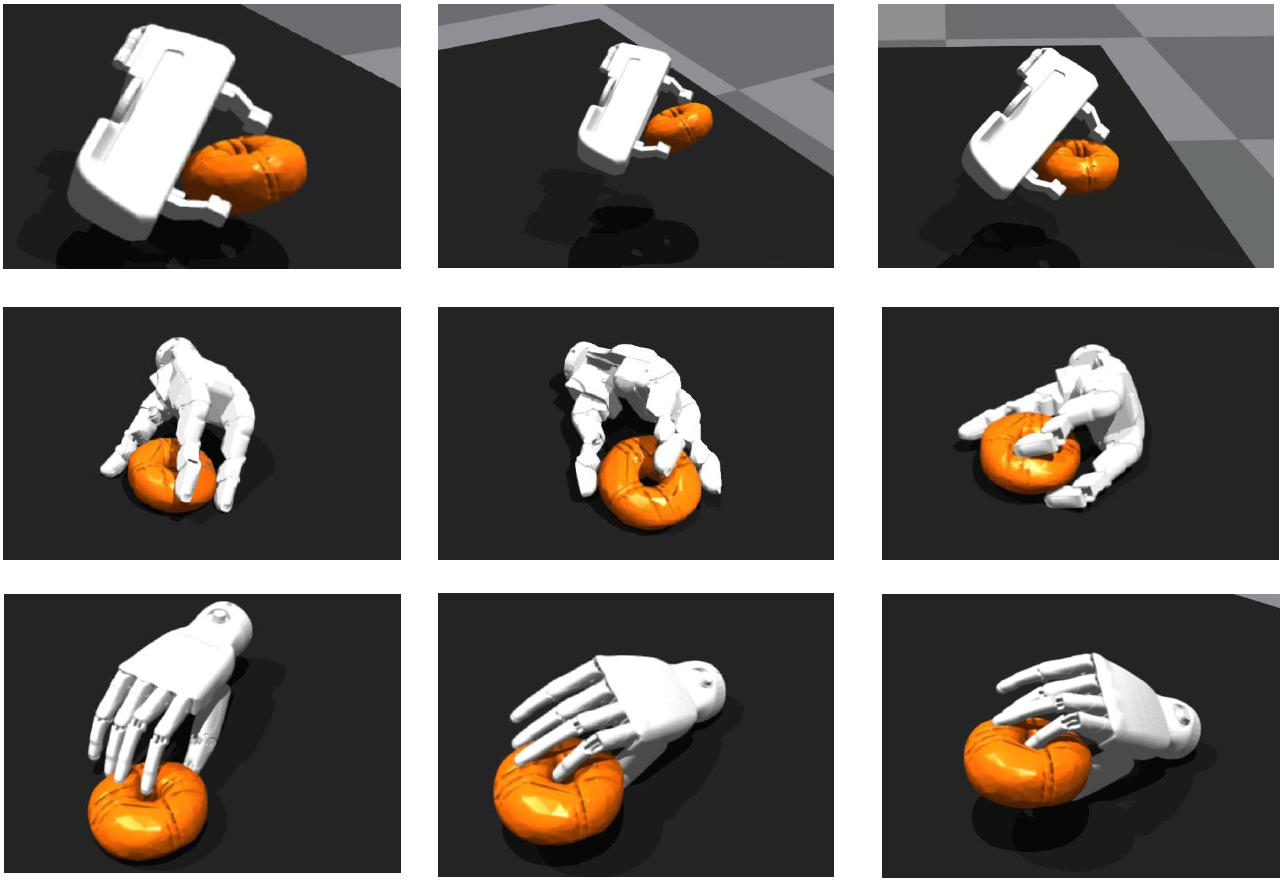}
   \caption{For the same task, UniBYD can learn different manipulation policies based on the physical characteristics of different robotic hands.}
   \label{fig:a1}
\end{figure}

As shown in \cref{fig:a1}, in this task, the two-fingered gripper directly grasps both sides of the donut. The three-fingered dexterous hand, having thicker fingers, uses its first and third fingers to grasp the sides while employing the middle finger to grip the donut from above.

In contrast, the more slender five-fingered dexterous hand uses its thumb to support the donut from below, while the middle and ring fingers grasp it from the sides. The most distinct difference compared to the 2- and 3-fingered hands is that UniBYD learned to pass the index finger through the center hole of the donut, thereby firmly securing it in coordination with the thumb.

\section{More Experimental Results}
\label{sec:me}
This section provides additional experimental results. Please refer to the video in the supplementary material for more detailed analysis and results.
\subsection{Ablation on Morphological Descriptors}
To evaluate the necessity of the hand-specific physical attributes, we compare UniBYD against a variant where the static morphological descriptor $v_{morph}^{h}$ is removed on a representative task. Omitting $v_{morph}^{h}$ leads to a significant performance degradation, with SR dropping from 93.00\% to 72.00\%, while OE and PE increase by 9\% and 62\%, respectively. 

Furthermore, $v_{morph}^{h}$ proves essential for training efficiency. UniBYD requires only ~80.7M steps to reach an 80\% SR, whereas the ablied version requires ~97.9M steps. This 18\% speedup confirms that providing explicit morphological information significantly reduces the complexity of policy search across diverse robotic embodiments.

\subsection{Hyperparameter Robustness Analysis}
To quantify robustness, we conduct a sensitivity sweep on critical parameters (SE Decay, Goal Reward Start Epoch, Entropy Decay) with a significant variation margin of $\pm 50\%$ on a representative task. As shown in \cref{tab:sensitivity}, even under extreme parameter deviations, the model maintains competitive performance ($>81\%$), demonstrating that it does not suffer from "cliff-edge" brittleness.
For parameters such as SE Decay, performance remains consistently near-perfect ($>93\%$) across the entire variation range.
This confirms that while reasonable curriculum scheduling aids training, the framework itself is robust to hyperparameter selection.

\begin{table}[h]
    \centering
    \caption{Hyperparameter Sensitivity Analysis. }
    \label{tab:sensitivity}
    \resizebox{1.0\linewidth}{!}{
        \begin{tabular}{l c c c} 
            \toprule
            \textbf{Hyperparameter} & 
            \textbf{Low Setting (-50\%)} & 
            \textbf{Default Setting} & 
            \textbf{High Setting (+50\%)} \\ 
            \midrule
            SE Decay Horizon ($T_{decay}$) & 
            93.59\% & 
            \textbf{99.83\%} & 
            96.17\% \\
            
            Goal Reward Start Epoch & 
            81.83\% & 
            \textbf{98.17\%} & 
            86.50\% \\
            
            Entropy Decay ($T_{entropy\_decay}$) & 
            84.50\% & 
            \textbf{98.50\%} & 
            98.17\% \\ 
            \bottomrule
        \end{tabular}
    }
\end{table}

\subsection{Component Ablation Analysis}
To validate the contribution of each module, we conduct an ablation study on a representative task as shown in \cref{tab:ablation_study}. Reward Annealing is critical. Removing it causes the sharpest performance drop (93.00\% $\to$ 75.44\%), indicating its vital role in bridging imitation and exploration. Boundary and Entropy Losses are essential for precision. Removing them notably degrades both SR and pose accuracy (higher OE/PE). The full UniBYD framework consistently outperforms all ablated variants, confirming the necessity of each component.

\begin{table}[htbp]
    \centering
    \caption{Ablation study of different components on a representative task.}
    \label{tab:ablation_study}
    \resizebox{\linewidth}{!}{ 
        \begin{tabular}{lcccc} 
            \toprule
            \textbf{Method} & \textbf{SR}$\uparrow$(\%) & \textbf{OE}$\downarrow$($^\circ$) & \textbf{PE}$\downarrow$(m) & \textbf{AS} \\ 
            \midrule
            w/o Boundary Loss    & 81.57 & 7.50 & 0.17 & 8.97 \\
            w/o Entropy Loss     & 84.68 & 5.67 & 0.18 & 8.34 \\
            w/o Reward Annealing & 75.44 & 5.80 & 0.20 & 7.65 \\
            UniBYD               & 93.00 & 2.92 & 0.10 & 9.12 \\
            \bottomrule
        \end{tabular}
    }
\end{table}

\subsection{Experimental Results in Simulation}
\begin{figure}[t]
  \centering
  \includegraphics[width=\linewidth]{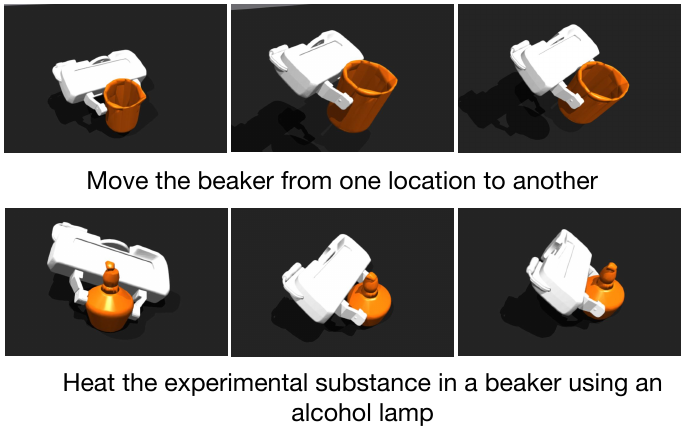}
   \caption{Experimental results of the 2-fingered robotic hand in simulation.}
   \label{fig:a2}
\end{figure}

\begin{figure}[t]
  \centering
  \includegraphics[width=\linewidth]{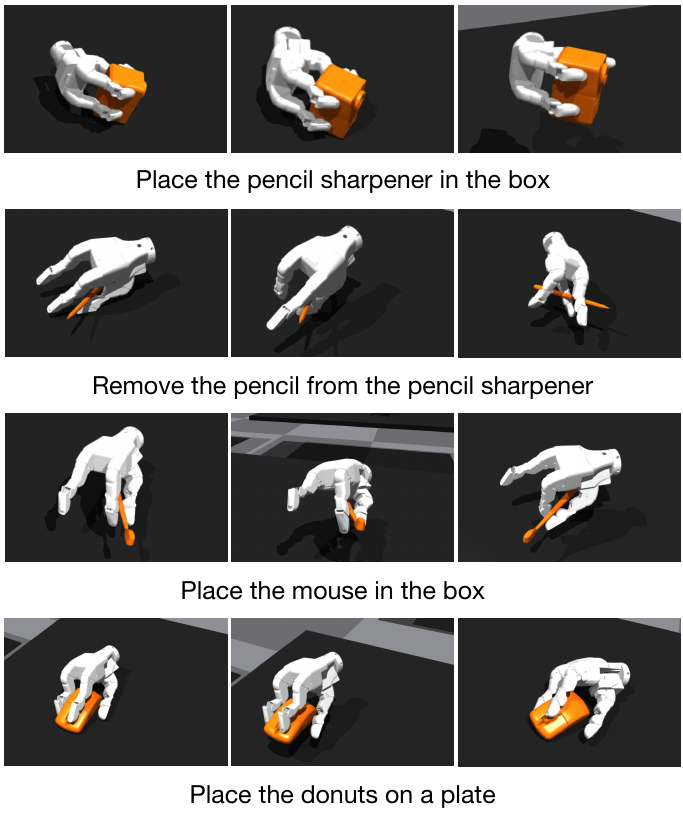}
   \caption{Experimental results of the 3-fingered robotic hand in simulation.}
   \label{fig:a3}
\end{figure}

\begin{figure}[t]
  \centering
  \includegraphics[width=\linewidth]{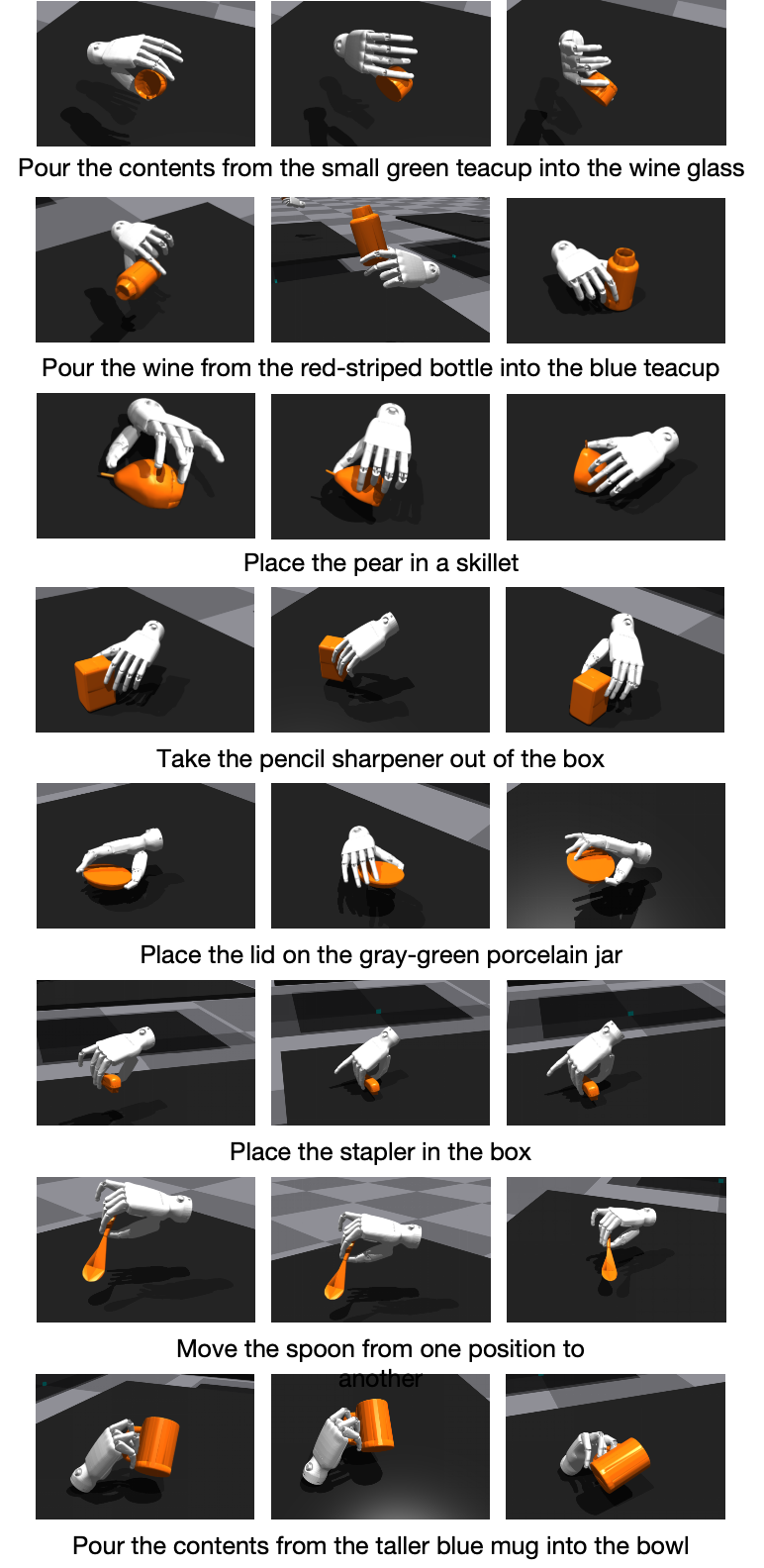}
   \caption{Experimental results of the single 5-fingered robotic hand in simulation.}
   \label{fig:a4}
\end{figure}

\begin{figure}[t]
  \centering
  \includegraphics[width=\linewidth]{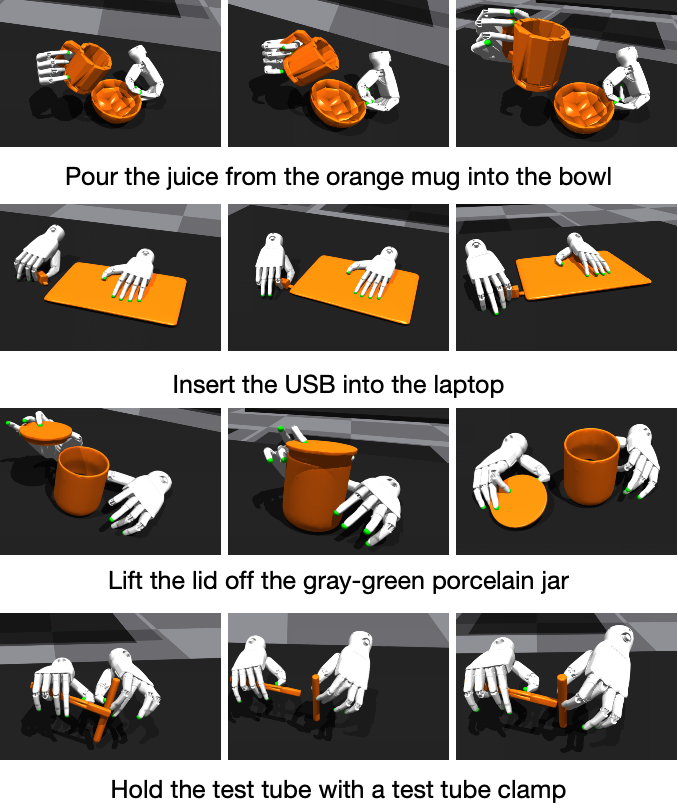}
   \caption{Experimental results of the dual 5-fingered robotic hand in simulation.}
   \label{fig:a5}
\end{figure}

This subsection presents the experimental results in simulation. The results for the 2-fingered, 3-fingered, single 5-fingered, and dual 5-fingered hands are shown in \cref{fig:a2}, \cref{fig:a3}, \cref{fig:a4}, and \cref{fig:a5}, respectively.

\subsection{Real-World Experimental Results}
\begin{figure}[t]
  \centering
  \includegraphics[width=\linewidth]{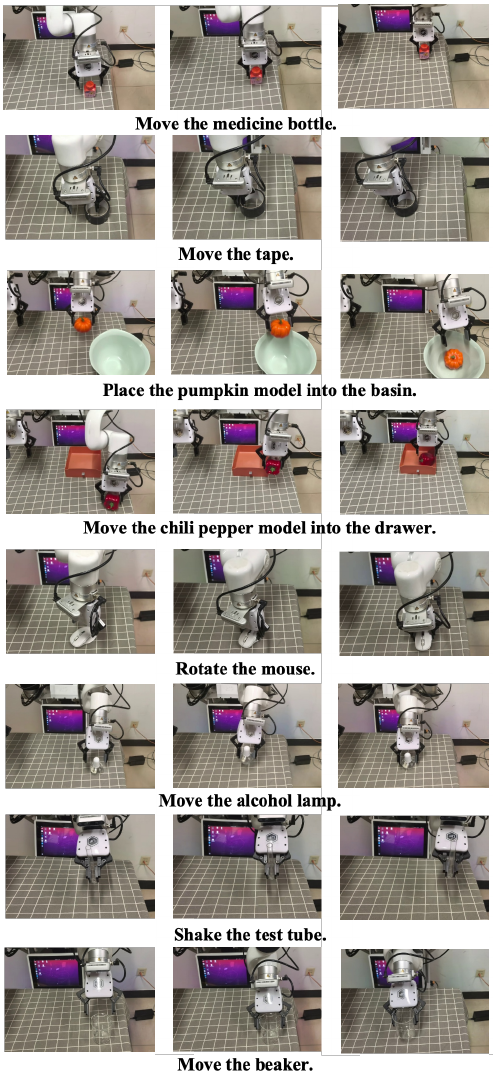}
   \caption{Experimental results of the 2-fingered robotic hand in the real world.}
   \label{fig:a6}
\end{figure}

\begin{figure}[t]
  \centering
  \includegraphics[width=\linewidth]{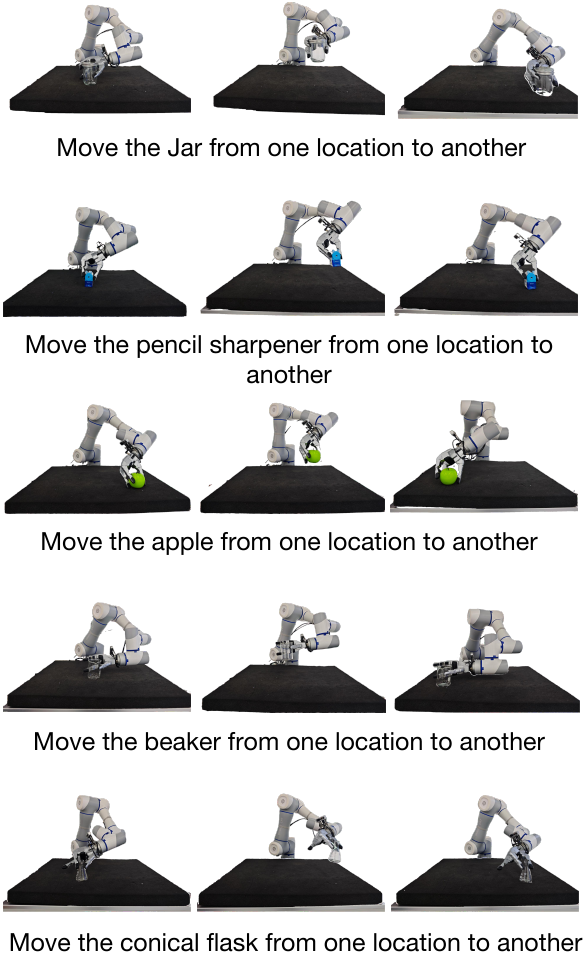}
   \caption{Experimental results of the 3-fingered robotic hand in the real world.}
   \label{fig:a7}
\end{figure}

\begin{figure}[t]
  \centering
  \includegraphics[width=\linewidth]{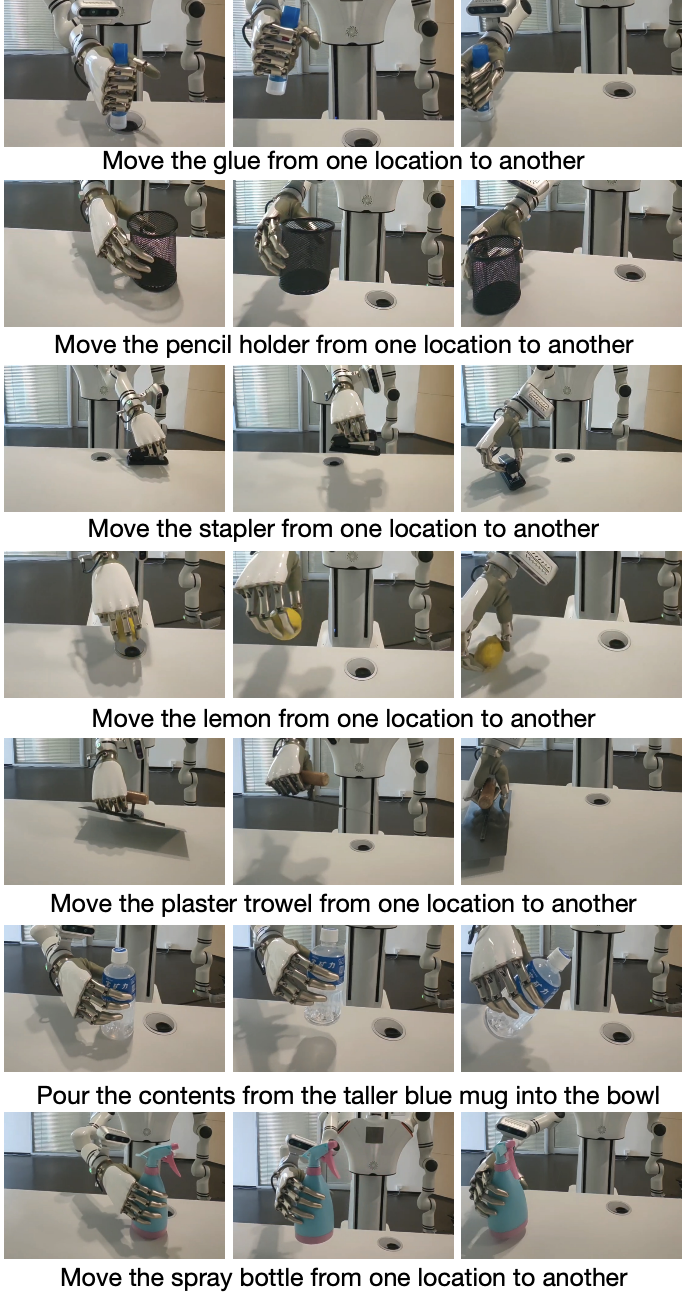}
   \caption{Experimental results of the 5-fingered robotic hand in the real world.}
   \label{fig:a8}
\end{figure}

This subsection presents the experimental results in the real world. The results for the 2-fingered, 3-fingered, and 5-fingered hands are shown in \cref{fig:a6}, \cref{fig:a7}, and \cref{fig:a8}, respectively.

\end{document}